\pdfoutput=1
\documentclass[11pt]{article}
\usepackage{acl}

\usepackage{titlesec}
\usepackage{times}
\usepackage{latexsym}
\usepackage[T1]{fontenc}
\usepackage[utf8]{inputenc}
\usepackage{microtype}
\usepackage{inconsolata}
\usepackage{arydshln}
\usepackage{graphicx}
\usepackage{tabularx,booktabs}

\usepackage{enumitem}
\usepackage{graphicx}
\usepackage{hyperref}
\usepackage{multirow}
\usepackage{graphicx}
\usepackage{tabularx}
\usepackage{url}
\usepackage{xcolor}
\usepackage{epsfig}
\usepackage{adjustbox}
\usepackage{amsfonts, amsmath, amssymb}
\usepackage{booktabs} 
\usepackage{comment}
\usepackage{caption, subcaption}
\usepackage{textcomp}
\usepackage{relsize}
\usepackage{stmaryrd}
\usepackage{bbm}
\usepackage{rotating}
\usepackage{helvet}
\usepackage{courier}
\usepackage{cleveref}
\usepackage{xspace}
\usepackage{enumitem}
\usepackage{epigraph}
\usepackage{mdframed}
\usepackage{makecell}
\usepackage{tcolorbox}
\usepackage{nicematrix}
\usepackage{tabu}
\usepackage{colortbl}
\usepackage{xcolor}
\PassOptionsToPackage{table}{xcolor}

\usepackage{pifont}

\usepackage{listings}
\newcolumntype{X}{>{\raggedright\arraybackslash}m{0.9\linewidth}}
\newcolumntype{W}{>{\centering\arraybackslash}m{0.06\linewidth}}
\newcolumntype{L}{>{\arraybackslash}m{0.99\linewidth}}

\title{\textsc{P-Check}: Advancing Personalized Reward Model \\ via Learning to Generate Dynamic Checklist}

\author{
    Kwangwook Seo~~~
    Dongha Lee\thanks{\; Corresponding author}\\
    Department of Artificial Intelligence,Yonsei University\\
    \texttt{\{tommy2130,donalee\}@yonsei.ac.kr}\\   
}

\begin{document}
\maketitle
\begin{abstract}
Recent approaches in personalized reward modeling have primarily focused on leveraging user interaction history to align model judgments with individual preferences.
However, existing approaches largely treat user context as a static or implicit conditioning signal, failing to capture the dynamic and multi-faceted nature of human judgment. 
In this paper, we propose \textbf{\textsc{P-Check}}, a novel personalized reward modeling framework, designed to train a plug-and-play checklist generator that synthesizes dynamic evaluation criteria for guiding the reward prediction.
To better align these checklists with personalized nuances, we introduce \textit{Preference-Contrastive Criterion Weighting}, a training strategy that assigns saliency scores to criteria based on their discriminative power for personalized judgment.
We conduct extensive experiments and demonstrate that \textsc{P-Check} not only improves reward accuracy but also enhances downstream personalized generation, and remains robust in OOD scenarios. \href{https://github.com/tommyEzreal/P-Check_}{[CODE]}
\end{abstract}

\section{Introduction}
\label{sec:introduction}
Recent advancements in Large Language Models (LLMs) have largely been driven by alignment techniques such as Reinforcement Learning from Human Feedback (RLHF), which rely on reward models to serve as proxies for human values and steer model behavior~\citep{ouyang2022traininglanguagemodelsfollow}. 
However, as LLMs are increasingly deployed as personalized assistants~\citep{xie2025a}, a reward model optimized on global preference distributions may not sufficiently address the diverse and subjective nature of individual user needs~\citep{li2025prefpalette}. 
This gap has led to a persistent demand for Personalized Reward Modeling, aiming to capture the intricate and subtle preferences that vary across users.
\begin{figure}[ht]
\centering
\includegraphics[width=\linewidth]{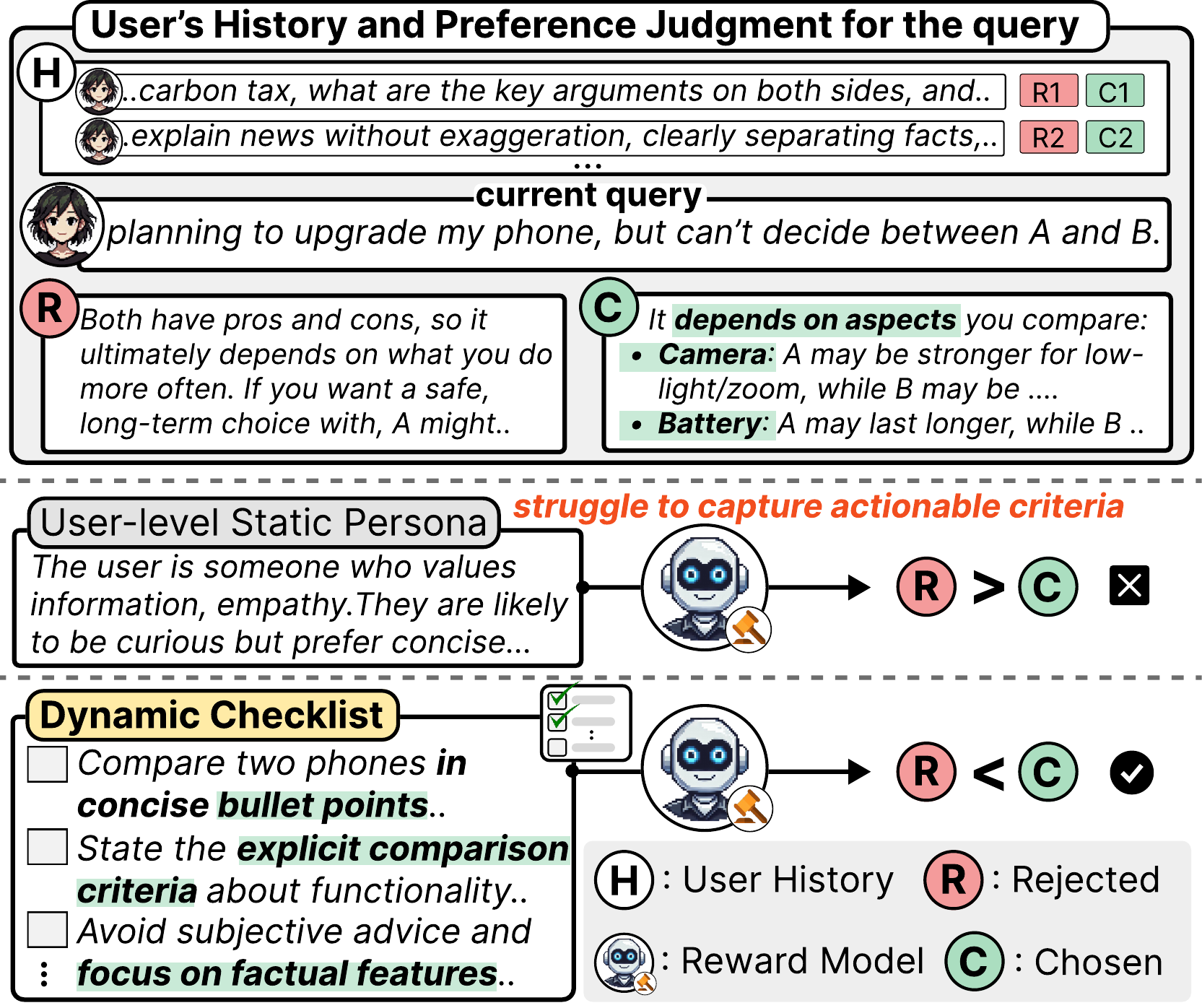}
% \caption{Motivating example illustrating human preference judgment. Beyond general profile, users form task-specific criteria \textit{on-the-fly} as the core evaluative factors shift, which determine the preferred response.}
\vspace{-0.75cm}
\caption{Motivating example of \textsc{P-Check}. Unlike the static persona that struggles to capture actionable criteria, dynamic checklist provides reliable guidance by explicitly specifying query-specific requirements.}
\vspace{-0.35cm}
\label{fig:motivation}
\vspace{-0.35cm}
\end{figure}

In response to these needs, existing works on personalized reward modeling~\citep{poddar2024personalizing,chen2025pal,ryan-etal-2025-synthesizeme} have primarily focused on leveraging user interaction history to align model judgments with individual preferences inferred from past interactions. 
While establishing an effective foundation, these approaches exhibit critical limitations in how they leverage user context for reward prediction.
\textbf{(1) Lack of explicit preference criterion:} user context is often expressed as a descriptive persona or an implicit conditioning signal, rather than an explicit guidance for evaluation. As a result, these representations provide limited support for modeling what concretely constitutes a user’s decision basis in a given judgment, which can hinder the accuracy and explainability of personalized reward prediction.
\textbf{(2) Static user-level preference signal:} since most existing approaches treat user context as a user-level static signal, they struggle to fully capture intra-user preference shifts that emerge across queries, where both the set of relevant evaluative factors and their relative importance can change with the task requested by the user. 

To address these limitations, we propose \textbf{\textsc{P-Check}}, a novel framework for personalized reward modeling that emulates the dynamic process of human preference judgment.
Our key idea is to augment a base reward model (\textit{i.e.}, LLM-as-a-Judge) with a plug-and-play checklist generator that learns to synthesize query-specific evaluation criteria from the interaction history of each user.
This is motivated by the insight that human evaluation is inherently multi-faceted~\citep{bakker2022finetuning, li2025prefpalette}, and the core evaluative factors shifts (\textit{e.g.}, Figure~\ref{fig:motivation}) depending on the task~\citep{hsee1999preference,pitis2024improving}. 
Humans can determine their preferred responses even in unfamiliar scenarios by forming these context-specific standards on-the-fly~\citep{gregory1993valuing, slovic1995construction}.
Reflecting this process, \textsc{P-Check} reframes personalized reward modeling to focus on learning explicit evaluation criteria behind the reward score, which gives more generalizable and transparent guidance for reward prediction.

One straightforward approach to training a checklist generator is to prompt strong LLMs to generate intermediate checklist from annotated preference pairs and distill it into a student model.
However, this naive distillation strategy encounters two key challenges: (1) \textbf{\textit{low clarity of personalized signal}}: annotated human preference pairs often mix objective errors with subjective dislikes~\citep{ziegler2020finetuninglanguagemodelshuman}, which may lead the generated checklists to merely evaluate generic quality instead of personalized criteria.
(2) \textbf{\textit{low discriminability of criteria}}: generated checklists often contain trivial or superficial criteria that dilute the discriminative signals necessary to distinguish core constraints from minor details in preference judgments.

To tackle these challenges, we introduce \textbf{Preference-Contrastive Criterion Weighting}, a novel checklist training strategy that assigns a personalized weight to each criterion based on its contribution to preference judgments. 
Specifically, to obtain informative personalized contrasts rather than generic quality signals, it first performs \textit{inter-user contrastive sampling}: in addition to the original rejected response, each preference pair is augmented with responses generated for other users with divergent preference axes. 
Then, \textit{personalized saliency scoring} calculates each criterion’s saliency by measuring the marginal drop in the relative separation between the chosen response and its contrastive set when that criterion is ablated from the checklist.
The resulting scores serve as additional supervision labels for training the checklist generator, which provides more discriminative and personalized signals to the reward prediction.

We conduct extensive experiments on three personalized reward benchmarks spanning both in-distribution and out-of-distribution (OOD) settings. Our evaluation mainly focuses on (1) helpfulness of \textsc{P-check} in assigning accurate reward and (2) the versatility of \textsc{P-Check} in improving personalized alignment. 
Across all benchmarks, \textsc{P-Check} consistently outperforms existing personalized reward models in reward accuracy and shows strong robustness in OOD scenarios. 
We further show that its checklist-based signals are effective both as reward inputs to popular alignment strategies (\textit{i.e.}, DPO and Best-of-$N$ selection) and as a verbal feedback that can be directly returned to the generator, enabling lightweight personalization without policy parameter updates. Our contributions are:
% We summarize our contributions as follows:
% \begin{itemize}
%     \item We propose \textsc{P-Check}, a novel personalized reward modeling framework designed to train a plug-and-play checklist generator that dynamically constructs evaluation criteria for guiding the personalized reward prediction.
    
%     \item To train a reliable checklist generator, we introduce Preference-Contrastive Criterion Weighting, which assigns weights to each checklist criteria based on their discriminative power for personalized judgment.
    
%     \item Extensive experiments on three personalized reward benchmarks shows that \textsc{P-Check} not only improves reward accuracy but also enhances downstream personalized generation, and remains robust in OOD scenarios.
% \end{itemize}
\begin{itemize}[leftmargin=*, itemsep=2pt, topsep=2pt, parsep=0pt, partopsep=0pt]
    \item We propose \textsc{P-Check}, a novel personalized reward modeling framework designed to train a plug-and-play checklist generator that dynamically constructs evaluation criteria for guiding the personalized reward prediction.
    \item To train a reliable checklist generator, we introduce Preference-Contrastive Criterion Weighting, which assigns weights to each checklist criterion based on their discriminative power for personalized judgment.
    \item Extensive experiments on three personalized reward benchmarks show that \textsc{P-Check} not only improves reward accuracy but also enhances downstream personalized generation, while remaining robust in OOD scenarios.
\end{itemize}

\section{Preliminary Analysis}
\label{sec:prelim}

Following the recent rubric-based evaluation works~\citep{gunjal2025rubricsrewardsreinforcementlearning, liu2025openrubricsscalablesyntheticrubric}, we define a checklist as a candidate-agnostic evaluation plan that specifies what to check for a given user history and query.
Under this definition, we conduct two preliminary analyses asking (1) whether current LLM-based judges can reliably infer the right evaluation criteria from user context, and (2) whether making those criteria explicit as a checklist can improve the personalized judgments.

For the analyses, we sample 100 users from the PersonalRewardBench~\citep{ryan-etal-2025-synthesizeme}, where each instance consists of a user’s interaction history, a current query, and a human-annotated chosen-rejected response pair. 
% We retain instances that provide at least four non-overlapping chosen–rejected pairs for the same user and query to conduct the experiment in the second analyses. 
Please refer to the Appendix~\ref{app:prelim_details} for detailed experimental setups.

\begin{figure}[t]
\centering
\includegraphics[width=\linewidth]{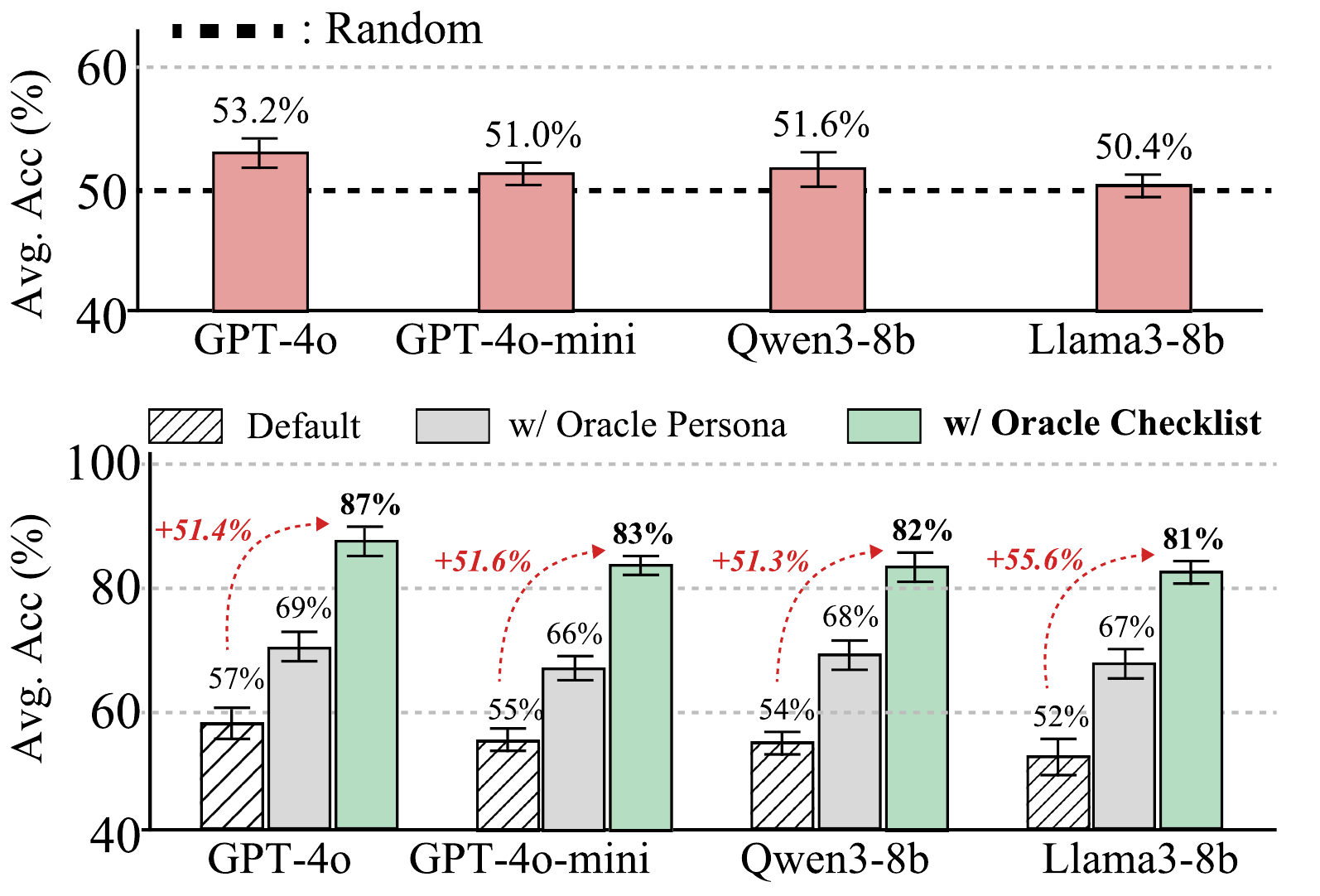}
\caption{\textbf{(Upper)} Analysis 1. \textbf{(Lower)} Analysis 2.}
\vspace{-0.3cm}
\label{fig:prelim}
\vspace{-0.35cm}
\end{figure}

\textbf{Analysis 1: LLMs Struggle to Infer Explicit Evaluation Criteria from User Context.}
We first test the current LLM's ability to infer the right evaluation criteria for a given user and query via a binary classification task. 
Given the user’s interaction history and the current query, the LLM is prompted to select the more appropriate option that can explain the user's judgment from (i) an oracle checklist and (ii) a counter-preference checklist.
We construct the oracle checklist to justify the chosen response, whereas the counter-preference checklist is constructed to justify the rejected one, by prompting an LLM with the user history, query, and labeled pair.
After generation, we conduct human verification to ensure both checklists are grounded in the user context and justify their respective responses.
Results in Figure~\ref{fig:prelim} (Upper) show that LLM-based judges perform near random guessing on identifying the oracle checklist, suggesting that they struggle to infer the reliable personalized criteria implied by the user context.

\textbf{Anaylsis 2: LLMs Make Better Personalized Judgements When Given an Explicit Checklist.}
We next assess whether LLM-based judges can select the personalized response when they are provided with an explicit checklist.
Given the user history, query, and the response pair, we prompt the LLM to choose the more personalized response with the checklist provided as additional input context.
Here, we evaluate preference selection on held-out pairs using oracle checklists derived from disjoint pairs for the same user and query.
As a baseline, we also provide an oracle persona generated only from the user history and labeled chosen–rejected pairs (with all samples verified by humans), but expressed as a descriptive user profile~\citep{ryan-etal-2025-synthesizeme} rather than explicit criteria.
Results in Figure~\ref{fig:prelim} (Lower) show that providing the checklist leads to a notable improvement in preference selection, while the persona yields only a marginal gain despite having access to the same user history.
These results suggest that explicit criteria provide more actionable guidance for preference selection than descriptive user personas, and highlight the necessity for dynamic use of user context with respect to the current query.

\section{\textsc{P-Check}: Proposed Method}
\label{sec:method}
\begin{figure*}[!ht]
\centering
\includegraphics[width=1\textwidth]{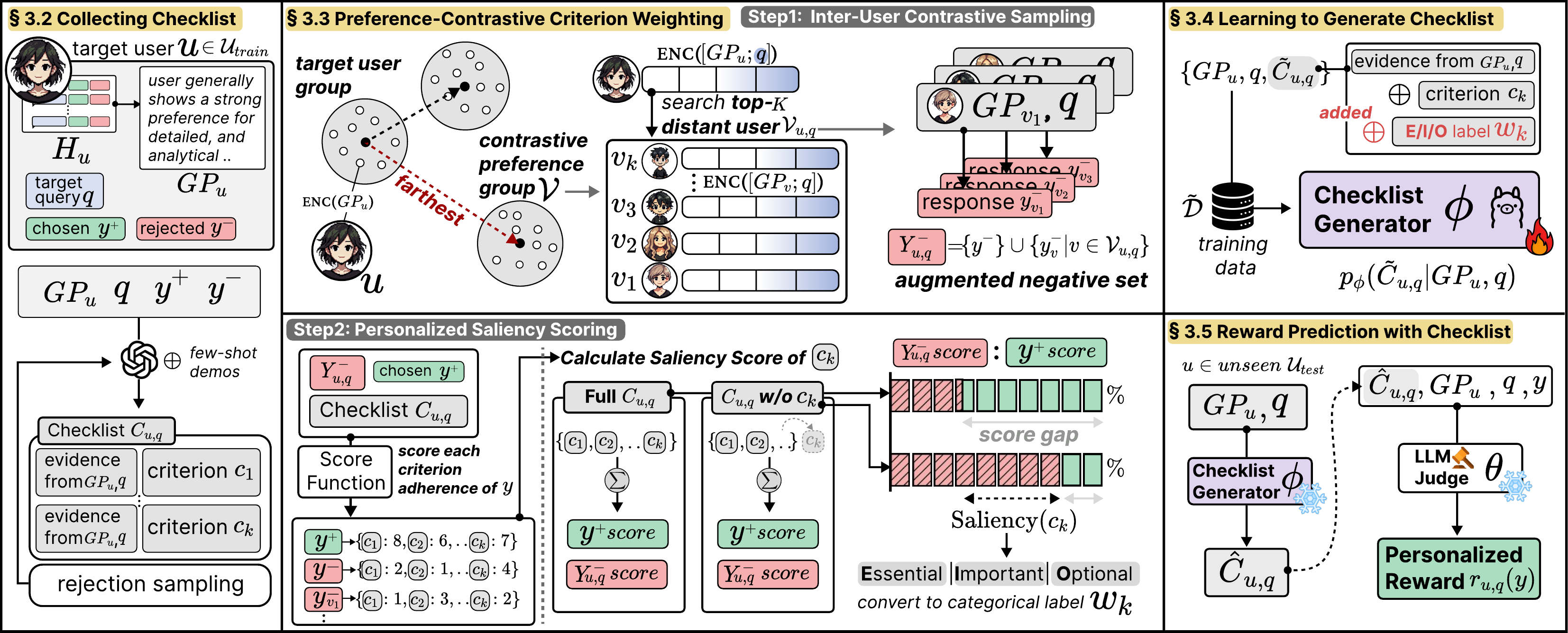}
\caption{Overview of \textsc{P-Check}, illustrating the training and inference processes.}
\vspace{-0.2cm}
\label{fig:method}
\vspace{-0.2cm}
\end{figure*}

Motivated by the insights in Section~\ref{sec:prelim}, we propose \textsc{P-Check}, a novel personalized reward modeling framework. 
We present an overview in Figure~\ref{fig:method}.

\subsection[Problem Formulation]{\hspace{-0.6em}Problem Formulation}
Recent approaches in personalized reward modeling~\citep{chen2025pal,ryan-etal-2025-synthesizeme} typically condition reward prediction on a user’s interaction history. 
In this setting, each user $u$ is associated with a history $H_u$ consisting of pairwise preference judgments over past queries and responses. 
Given a current query $q$, user history $H_u$, and a candidate response $y$, the task is to predict a reward $r$ that reflects how well $y$ satisfies the user’s personalized intent implied by $H_u$. 
While the end task is reward prediction, our focus is on learning to generate a personalized checklist $C_{u,q}$ as an intermediate representation of the user’s decision basis and using it as additional input to guide the reward model $\theta$:
\begin{equation}
r \sim P_{\theta}\!\left(\cdot \mid y, q, H_u, C_{u,q}\right)
\end{equation}

\subsection[Collecting Checklist from Preference Data]{\hspace{-0.6em}Collecting Checklist from Preference Data}
To train the checklist generator, we collect synthetic checklists from logged pairwise preferences in the interaction histories of users in the training split.
Specifically, we start from a preference dataset $\mathcal{D}=\{(H_u, q, y^{+}, y^{-})\}$, where $H_u$ is the user’s history of pairwise preference judgments and $(q, y^{+}, y^{-}) \notin H_u$ is the target preference instance. 
First, we generate a compact user-level summary $GP_u$ from $H_u$, encouraging checklist to reflect general preference patterns of each user rather than instance-specific details in each interaction history. 
Using this $GP_u$ and $q$ as context, we prompt an LLM to synthesize a checklist that contrasts the chosen and rejected responses $(y^+, y^-)$ while grounding its output in evidence implied by $GP_u$ and $q$. 
Through this prompt, we obtain a  checklist $C_{u,q}$ consisting of multiple decision-relevant criteria $\{c_k\}$ that pass on $y^+$ but fail on $y^-$.
\begin{equation}
C_{u,q} \;=\; \mathrm{LLM}\!\left(GP_u,\; q,\; y^{+},\; y^{-}\right)
\end{equation}

\subsection[Preference-Contrastive Criterion Weighting]{\hspace{-0.6em}Preference-Contrastive Criterion Weighting}
Based on the synthetic checklists $C_{u,q}$, the most straightforward approach is to directly supervise the checklist generator to synthesize these criteria.
However, simply optimizing for these raw checklist sequences carries some potential downsides. 
Specifically, the generated checklists may entangle objective validity with subjective preference, causing personalized signals to be confounded by generic quality, which makes it difficult to capture the nuances of personal intent. 
Furthermore, they often contain trivial items which dilute the decision-critical signal, resulting in low discriminability. 

To mitigate the aforementioned bottlenecks in training a checklist generator, we introduce \textit{Preference-Contrastive Criterion Weighting}, which first samples preference-relevant contrasts to obtain rich personalization signal, and assigns weight to each criterion $c_k$ in a $C_{u,q}$ that reflects its contribution to the preference decision.
We formulate this process in two main steps:

% \paragraph{Step1: Inter-User Contrastive Sampling.}
\noindent \textbf{Step 1: Inter-User Contrastive Sampling.}
To obtain preference-relevant signals, we sample additional negatives from users with distant preference directions from the target user, so that the augmented negative pool highlights contrasts along the target user’s preference axes.
To efficiently identify such distant users without exhaustive pairwise comparisons, we employ a two-stage retrieval process.
First, we perform coarse-grained filtering by clustering users based on their general preference $GP_u$.
For a target user $u$, we identify the cluster centroid farthest from $u$'s cluster and sample a candidate set of contrastive preference users $\mathcal{V}$ from that distant cluster.
Next, to account for query-dependent preference shifts, we perform fine-grained selection.
We compute query-conditioned embeddings for the target user and each candidate $v\in\mathcal{V}$ by jointly encoding their respective general preferences with the target query $q$ (\textit{i.e.}, $\text{Enc}(GP_x, q)$ for $x\in\{u\}\cup \mathcal V$).
% We then compute query-conditioned representations for the $u$ and candidates $v\in \mathcal V$ by jointly encoding each $GP_x$ with $q$ (for $x\in\{u\}\cup \mathcal V$), 
This keeps distances anchored in user-level preference signals encoded in $GP_x$ while remaining sensitive to query-level preference shifts.
From these embeddings, we select the top-$K$ most distant users $\mathcal{V}_{u,q} \subseteq \mathcal{V}$. Finally, we prompt an LLM to generate responses $y^-_v$ for each $v \in \mathcal{V}_{u,q}$ from $(GP_v, q)$, constructing a diverse negative pool that delineates the target user's personal preference axes.
% From these embeddings, we select the top-$K$ distant candidates $\mathcal{V}_{u,q}\subseteq \mathcal{V}$, and prompt an LLM to generate counter-preference responses $y^-_{v}$ from $(GP_v, q)$ for each $v\in \mathcal{V}_{u,q}$, which augments the negative pool.

% \paragraph{Step2: Personalized Saliency Scoring.}
\noindent \textbf{Step2: Personalized Saliency Scoring.}
Building on the augmented negative set $Y^-_{u,q}=\{y^{-}\}\cup\{y^{-}_{v} | v\in \mathcal{V}_{u,q}\}$ constructed from step 1, the goal of this step is to compute the weight of each criterion $c_k$ in checklist $C_{u,q}$ by evaluating how consistently it separates the chosen response from these negatives. 
Since these synthetic negatives are not user-verified rejection labels, we do not use them to propose new criteria.
Instead, we keep the criterion set fixed to $C_{u,q}$ and use the additional negatives only as evidence for weighting.
Specifically, we define the saliency of a criterion $c_k$ as the marginal decrease in the relative checklist adherence gap between the chosen response and the negative pool when $c_k$ is ablated from the checklist.
This definition follows the intuition that \textit{removing an important criterion makes the negatives appear relatively closer to the chosen response}.
To score checklist adherence, we use an LLM as a scoring function $f(C_{u,q},y)$ that outputs a criterion-wise score vector, where each entry $f_k(C_{u,q},y)$ is a 1-10 scale scalar score for a criterion $c_k\in C_{u,q}$. We aggregate these criterion-wise scores into a response-level checklist score $s(C_{u,q},y)$, and summarize the negative pool by averaging this score over $y\in Y^-_{u,q}$:

\begin{equation}
\small
s(C_{u,q},y)=\sum_{c_k\in C_{u,q}} f_k(C_{u,q},y)
\end{equation}

\begin{equation}
\small
s(C_{u,q},Y^-_{u,q})=\frac{1}{|Y^-_{u,q}|}\sum_{y\in Y^-_{u,q}} s(C_{u,q},y)
\end{equation}
We denote the checklist with $c_k$ removed as $C^{-k}_{u,q}$, and compute saliency by comparing the normalized negative-to-chosen ratio before and after ablation:
\begin{equation}
\small
\mathrm{Saliency}(c_k)=
\frac{s(C^{-k}_{u,q},Y^-_{u,q})}{s(C^{-k}_{u,q},y^{+})+\epsilon}
-
\frac{s(C_{u,q},Y^-_{u,q})}{s(C_{u,q},y^{+})+\epsilon}
\end{equation}
where $\epsilon$ is a small constant for numerical stability, and since 
$f$ returns a score vector, all $s(C^{-k}_{u,q},y)$ for $C_{u,q}$ can be computed from only a single scoring pass by excluding the $k$-th entry from the same vector.
Intuitively, we use a ratio that normalizes the checklist score of the negative pool by the checklist score of the chosen response, which can be read as ``\textit{how much the negatives catch up to the chosen response}'' under the checklist. Based on this ratio, saliency measures how much this 'catch up' increases when $c_k$ is removed, meaning that $c_k$ helps keep negatives apart from the chosen.

After computing the saliency scores, we apply ReLU to obtain a non-negative saliency signal that is easier to interpret. The resulting values are then rescaled within each checklist to obtain a distribution of criterion weights that sum to one, capturing relative importance across criteria.

\subsection[Learning to Synthesize Dynamic Checklist]{\hspace{-0.6em}Learning to Synthesize Dynamic Checklist}
With the annotated checklist $C_{u,q}$ and a criterion weight assigned to each $c_k$, our goal is to train a checklist generator $\phi$ that generates a query-level checklist from $(GP_u, q)$.
To align the supervision signal with our LM-based generator $\phi$, we first verbalize the continuous criterion weights into discrete natural language labels $w_k$ (\texttt{Essential}, \texttt{Important}, and \texttt{Optional}, denoted as \textbf{\texttt{E/I/O}}), allowing $\phi$ to seamlessly learn them as part of its output sequence. 
Specifically, we sort criteria in $C_{u,q}$ by their weights in descending order and traverse the ranked list while tracking the cumulative sum of weights. 
Using two thresholds $\tau_1$ and $\tau_2$, a criterion is labeled \textbf{\texttt{E}} if it falls within the top $\tau_1$\% of the cumulative sum, \textbf{\texttt{I}} if it falls between $\tau_1$\% and $\tau_2$\%, and \textbf{\texttt{O}} otherwise.

Through this process, we construct a checklist training set $\tilde{\mathcal{D}}=\{(GP_u, q, \tilde{C}_{u,q})\}$, where each $\tilde{C}_{u,q}$ includes criteria $c_k$ and their supporting evidences grounded in $(GP_u, q)$, augmented with the verbalized weight labels $w_k\in\{\textbf{\texttt{E}},\textbf{\texttt{I}},\textbf{\texttt{O}}\}$, so that the entire checklist can be learned as a single target sequence.
We then train the checklist generator $\phi$ on $\tilde{\mathcal{D}}$ with a standard next-token prediction objective to generate $\tilde{C}_{u,q}$ conditioned on $(GP_u, q)$:
\begin{equation}
\small
\mathcal{L}_{\phi}
=
-\sum_{(GP_u,q,\tilde{C}_{u,q})\in\tilde{\mathcal{D}}}
\log p_{\phi}\!\left(\tilde{C}_{u,q}\mid GP_u, q\right)
\end{equation}

\subsection[Personalized Rewarding with Checklist]{\hspace{-0.5em}Personalized Rewarding with Checklist}
At inference time, we use the trained checklist generator $\phi$ to infer a checklist $\hat{C}_{u,q}=\phi(GP_u, q)$ that guides an off-the-shelf reward model $\theta$ to predict the personalized reward $r_{u,q}(y)$ for a candidate $y$.
Here, $\theta$ is implemented as an LLM-Judge~\citep{zheng2023judging} that takes $(GP_u,q,y,\hat{C}_{u,q})$ as input and outputs a criterion-wise score (1-10 scale) vector.
Each item in $\hat{C}_{u,q}$ consists of criterion $c_k$ and its categorical label $w_k\in\{\texttt{E},\texttt{I},\texttt{O}\}$ indicating the relative saliency.
We map each label $w_k$ to a numerical weight based on the validation accuracy, which forms the weight vector $\mathbf{w}_{u,q}$.
The final scalar reward is obtained by the dot product between $\mathbf{w}_{u,q}$ and the criterion-wise score vector inferred from $\theta$:
\begin{equation}
r_{u,q}(y)=\mathbf{w}_{u,q}^{\top}\,\theta(GP_u,q,y,\hat{C}_{u,q})
\end{equation}

% It is worth noting that this relative ratio focuses on separation and is less affected by overall score shifts induced by checklist ablation, yielding a more robust importance signal.

% It is worth noting that this relative ratio focuses on separation and is less affected by overall score shifts that can occur when a criterion is removed from the checklist for ablation.
% This normalization yields a more robust importance signal by emphasizing relative separation rather than absolute score magnitudes. 
% used in prior works~\citep{pmlr-v70-sundararajan17a,Lundberg2017AUA}.
% Finally, we clip negative saliency values to zero (ReLU) and normalize $\{\mathrm{Saliency}(c_k)\}$ across criteria to obtain instance-specific weights $\{w_k\}$, which are used as importance supervision in training the checklist generator.

\section{Experiments}
\label{sec:experiments}
To demonstrate the effectiveness of \textsc{P-Check}, we conduct extensive experiments across both in-distribution (ID) and out-of-distribution (OOD) benchmarks for personalization, focusing on its (1) helpfulness in predicting accurate reward and the (2) versatility in improving personalized alignment.
\subsection{Experimental Setups}

\newcommand{\pmv}[2]{#1\,{\scriptsize\color{black!45}{$\pm$\,#2\%}}}

\begin{table*}[t]
\centering
\small
\setlength{\tabcolsep}{4.5pt}
\renewcommand{\arraystretch}{1.1}

\begin{tabular}{lcccccc|c}
\toprule
\multirow{2.5}{*}{\textbf{Method}}
& \multicolumn{2}{c}{\textbf{PRISM-Personal. (ID)}}
& \multicolumn{2}{c}{\textbf{ARENA-Personal. (OOD)}}
& \multicolumn{2}{c|}{\textbf{BESPOKE-Meta. (OOD)}}
& \multirow{2.5}{*}{\textbf{AVG.}} \\
\cmidrule(lr){2-3}\cmidrule(lr){4-5}\cmidrule(lr){6-7}
& \textbf{Llama3-8b} & \textbf{Llama3-3b}
& \textbf{Llama3-8b} & \textbf{Llama3-3b}
& \textbf{Llama3-8b} & \textbf{Llama3-3b}
& \\
\midrule

\rowcolor{gray!12}
\multicolumn{8}{l}{\textit{Finetuned Reward Model}} \\
GPO          & \pmv{56.48}{1.74} & \pmv{55.26}{1.61} & \pmv{52.01}{3.26} & \pmv{51.89}{3.14} & \pmv{51.49}{1.85} & \pmv{52.04}{1.93} & 53.20  \\
VPL          & \pmv{58.23}{2.86} & \pmv{58.26}{2.23} & \pmv{53.36}{3.17} & \pmv{52.34}{3.28} & \pmv{53.54}{2.19} & \pmv{52.89}{2.79} & 54.77 \\
PAL          & \pmv{54.23}{1.85} & \pmv{56.81}{2.18} & \pmv{53.89}{3.15} & \pmv{52.41}{3.65} & \pmv{51.33}{2.30} & \pmv{51.49}{2.56} & 53.36 \\
BT + SynthMe & \pmv{\underline{62.74}}{1.79} & \pmv{\textbf{61.39}}{1.39} & \pmv{56.42}{3.44} & \pmv{53.32}{3.19} & \pmv{50.47}{2.16} & \pmv{50.83}{2.01} & \underline{55.86} \\
\midrule

\rowcolor{gray!12}
\multicolumn{8}{l}{\textit{In context LLM-as-a-Judge}} \\
Default       & \pmv{52.80}{2.57}  & \pmv{51.65}{2.37} & \pmv{53.56}{4.16} & \pmv{52.23}{4.13} & \pmv{55.46}{3.04} & \pmv{53.45}{3.13} & 53.19 \\
+ Memory      & \pmv{54.17}{1.40} & \pmv{50.86}{1.61} & \pmv{58.15}{4.33} & \pmv{52.29}{4.10} & \pmv{57.75}{2.87} & \pmv{54.23}{3.41} & 54.58 \\
+ CoT \textit{distill}        & \pmv{55.47}{1.90} & \pmv{53.36}{2.50} & \pmv{55.84}{2.96} & \pmv{\underline{53.77}}{3.05} & \pmv{\underline{61.35}}{1.87} & \pmv{\underline{55.23}}{1.95} & 55.84 \\
+ SynthMe     & \pmv{55.24}{1.71} & \pmv{52.09}{1.61} & \pmv{\underline{58.83}}{3.87} & \pmv{52.76}{3.65} & \pmv{54.67}{2.83} & \pmv{51.35}{2.70} &  54.16\\
+ \textbf{\textsc{P-Check} (ours)}
              & \pmv{\textbf{65.11}}{1.44} & \pmv{\underline{60.91}}{1.39}
              & \pmv{\textbf{61.56}}{2.85} & \pmv{\textbf{55.03}}{3.17}
              & \pmv{\textbf{75.48}}{2.27} & \pmv{\textbf{62.43}}{2.36}
              & \textbf{63.62} \\
\bottomrule

\end{tabular}
\vspace{-0.25cm}

\caption{Evaluation results on personalized reward modeling. Following~\citep{ryan-etal-2025-synthesizeme}, we report the binary preference prediction accuracy on unseen users across ID (PRISM) and OOD (ARENA, BESPOKE) benchmarks. We employ Llama-3-8B-It. and 3B-It. for reward model $\theta$, with results reported over five runs ($\pm$ 95\% CI).}
\vspace{-0.4cm}

\label{table:main}
\end{table*}

\noindent \textbf{Datasets.}
We conduct experiments on three popular personalized reward benchmarks.
In particular, we employ {PRISM}-Personalized~\citep{ryan-etal-2025-synthesizeme} as the ID dataset, while adopting {ChatbotArena-Personalized} and BESPOKE-MetaEval~\citep{kim2025bespokebenchmarksearchaugmentedlarge} as the OOD datasets.
For all datasets, we apply a strict user-level split, ensuring that evaluation is performed solely on unseen test users.

\noindent \textbf{Implementation Details.}
We employ Llama-3.2-3B-Instruct as the backbone for our checklist generator $\phi$.
For the generation of $GP_u$ for all users and the initial checklist training data $C_{u,q}$, we use GPT-4o-mini, and apply rejection sampling~\citep{yuan2023scalingrelationshiplearningmathematical} to refine the quality of $GP_u$ and $C_{u,q}$.
In the inter-user contrastive sampling stage, we select the top-3 most distant users based on user representations extracted from Qwen3-Embedding-0.6B.
To verbalize the saliency scores into numerical weights, we define threshold values as $(\tau_1, \tau_2) = (0.4, 0.9)$.
We provide further training details and ablation studies regarding these hyperparameters in Appendix~\ref{app:implement_detail}.

\noindent \textbf{Baselines.}
% \paragraph{Baselines.}
We compare \textsc{P-check} against diverse baselines (implementation details of baselines are in Appendix~\ref{app:baselines}). 
(1) \textit{In-context LLM judges:}
For the \textbf{Default} setup, we simply prompt an LLM to select the better personalized response.
We also implement \textbf{Memory}, which augments the judge with retrieved user interaction data for each query, and \textbf{SynthMe}~\citep{ryan-etal-2025-synthesizeme}, which optimizes a user persona and demonstrations for preference selection.
For \textbf{CoT \textit{distill}}, we use the same user information as \textsc{P-check}, but generate training data where the output is a free-form rationale that justifies the labeled pair.
(2) \textit{Fine-tuned reward models:}
We include existing fine-tuned personalized reward models including \textbf{GPO}~\citep{zhao2024group}, \textbf{VPL}~\citep{poddar2024personalizing}, \textbf{PAL}~\citep{chen2025pal}, and Bradley-Terry \textbf{(BT)} model augmented with \textbf{SynthMe}.
For a fair comparison, all baselines (except Default and Memory) and the checklist generator $\phi$ of \textsc{P-check} are optimized only on \textsc{PRISM} train users, and directly evaluated on unseen test users for both ID and OOD benchmarks.

\subsection{Results and Analysis}
\paragraph{\textsc{P-Check} helps LLM-as-a-Judge in predicting personalized reward.}
We compare \textsc{P-Check} against a diverse set of personalized reward modeling baselines, including both fine-tuned reward models and context-augmented LLM-judge approaches.
As shown in Table~\ref{table:main}, incorporating \textsc{P-Check} into the off-the-shelf LLM-Judge outperforms all baselines across different model scales.
Specifically, \textsc{P-Check} achieves an average accuracy of 63.62\%, marking a substantial improvement of +19.61\% over the Default setting (53.19\%).
These results suggest that \textsc{P-Check} effectively models the user’s decision basis as explicit evaluation criteria, which provides more actionable guidance for personalized reward prediction.
\paragraph{\textsc{P-Check} effectively generalizes across OOD benchmarks.}
Consistent with previous findings~\citep{zhang2025bradleyterrymultiobjectiverewardmodeling}, we observe that fine-tuned scalar reward models still struggle to generalize in reward prediction, often failing to transfer their performance to unseen user distributions.
In contrast, as shown in Table~\ref{table:main}, \textsc{P-Check} maintains robust performance in OOD scenarios and  outperforms all baselines.
We attribute this robustness to \textsc{P-Check}'s focus on learning the intrinsic preference logic via dynamically generating and weighting personal criteria, which equips the model with the ability to infer decision factors on-the-fly.

% (preamble 필요)
% \usepackage{booktabs,tabularx,array}
% \usepackage{amsmath} % optional

\begin{table*}[t]
\centering
\small
\setlength{\tabcolsep}{2pt}
\renewcommand{\arraystretch}{1.0}

\newcolumntype{Y}{>{\centering\arraybackslash}X}
\newcolumntype{P}{>{\raggedright\arraybackslash}p{0.17\textwidth}}

% best / second-best formatting
\newcommand{\best}[1]{\textbf{#1}}
\newcommand{\second}[1]{\underline{#1}}

% significance mark (0-width overlay so alignment doesn't shift)
\newcommand{\sig}{\rlap{$^{*}$}} % or: \newcommand{\sig}{\rlap{\,$^{*}$}}

\begin{tabularx}{\textwidth}{P YYY YYY YYY}
\toprule
\textbf{Align. Method}
& \multicolumn{6}{c}{\textbf{\textit{Best-of-N Selection (BoN)}}}
& \multicolumn{3}{c}{\textbf{\textit{DPO}}} \\
\textbf{Policy $\pi$}
& \multicolumn{3}{c}{\textbf{Llama3-8b}}
& \multicolumn{3}{c}{\textbf{GPT-4o-mini}}
& \multicolumn{3}{c}{\textbf{Llama3-8b}} \\
\cmidrule(lr){2-4}\cmidrule(lr){5-7}\cmidrule(lr){8-10}
\textbf{RM($\downarrow$) / Metric($\rightarrow$)}
& R-L & METEOR & B.Eval
& R-L & METEOR & B.Eval
& R-L & METEOR & B.Eval \\
\midrule

Default $\pi$
& 7.92 & 6.09 & 51.55
& 7.83 & 6.43 & 51.69
& 7.92 & 6.09 & 51.55 \\

+ VPL
& 8.05 & 7.06 & 52.19
& 8.45 & 7.49 & 53.36
& 8.68 & 7.70 & 53.55 \\

+ PAL
& 8.32 & 7.10 & 51.68
& 8.31 & 7.46 & 52.03
& 7.42 & 6.11 & 53.49 \\

+ CoT \textit{distill}
& 8.24 & 7.14 & 54.90
& 8.24 & 7.62 & 54.76
& 8.41 & 7.63 & 56.64 \\

+ SynthMe
& 7.99 & 6.85 & 52.06
& 8.01 & 6.46 & 52.96
& 7.74 & 6.45 & 52.46 \\

+ \textbf{\textsc{P-Check} (ours)}
& \best{9.43}\sig & \best{8.22}\sig & \best{59.76}\sig
& \best{9.61}\sig & \best{8.47}\sig & \best{57.32}\sig
& \best{9.94}\sig & \best{9.67}\sig & \best{61.21}\sig \\
\bottomrule
\end{tabularx}
\vspace{-0.25cm}

\caption{Evaluation results on personalized generation in BESPOKE using various reward models. Statistical significance(*) is assessed using a paired t-test over five runs against the strongest baseline in each setting ($p<0.05$).}
\vspace{-0.3cm}

\label{table:align}
\end{table*}

\begin{figure}[ht]
\centering
\includegraphics[width=\linewidth]{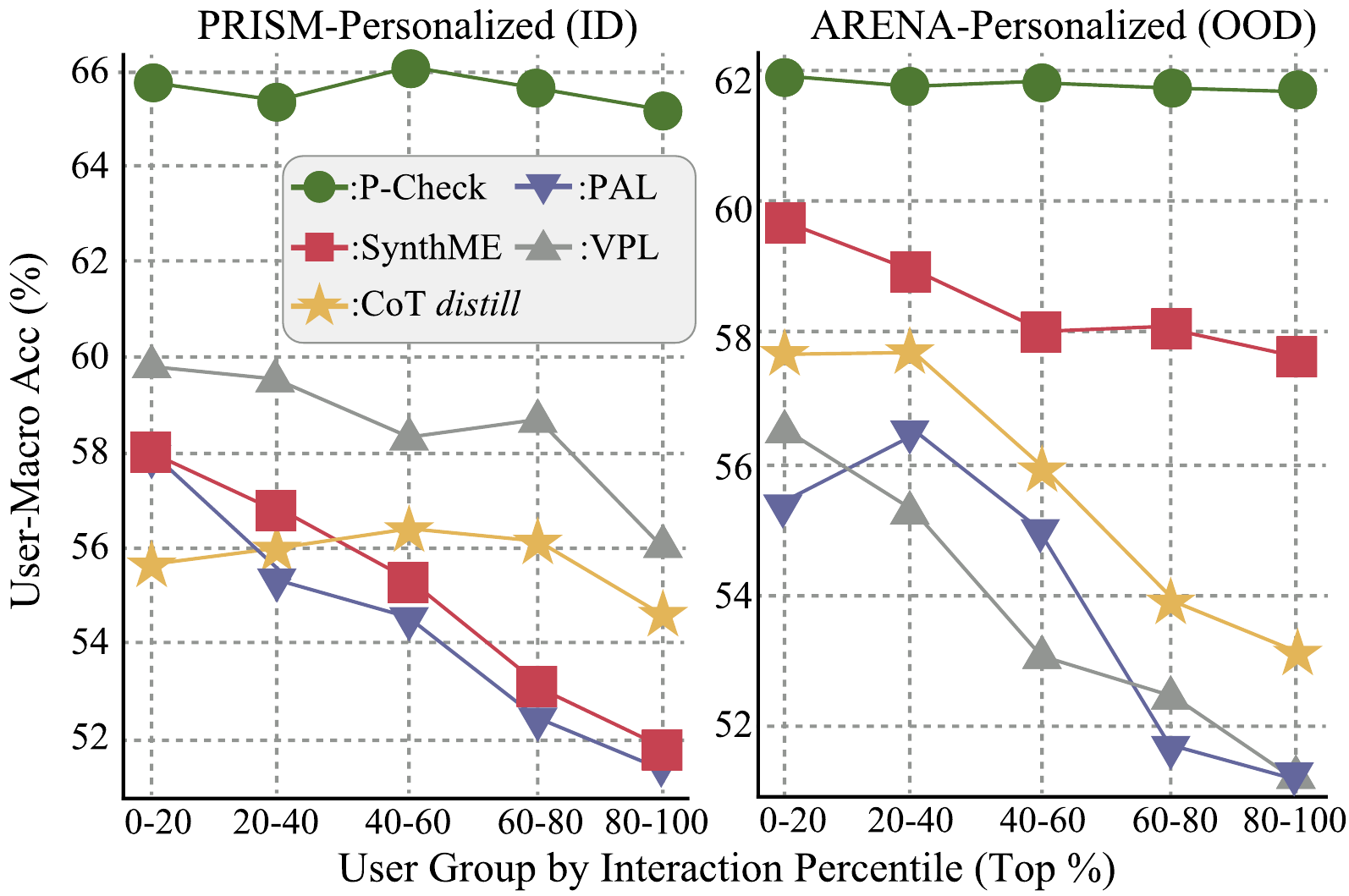}
\vspace{-0.7cm}
\caption{Experiments on sparse interaction scenario.}
\label{fig:sparse}
\vspace{-0.5cm}
\end{figure}

\paragraph{\textsc{P-Check} shows robustness in sparse user interaction scenarios.}
Real-world personalization often faces the challenge of long-tail users with limited interaction data. 
To evaluate whether \textsc{P-Check} remains robust in this real-world scenario, we examine the reliability of reward prediction under sparse test-time histories.
Specifically, we categorize test users into interaction-count percentiles, where higher percentiles correspond to increasingly sparse observed histories. 
We report User-Macro Accuracy for each bucket on PRISM (ID) and ARENA (OOD) to decouple performance from the number of test pairs per user.
As shown in Figure~\ref{fig:sparse}, while baseline methods exhibit consistent degradation as interaction history becomes scarcer, \textsc{P-Check} remains relatively stable across all user groups in both benchmarks.
These results suggest that while baselines struggle to recover global preference signals from sparse interactions with high estimation uncertainty, \textsc{P-Check} mitigates this degradation by learning to synthesize only query-specific comparison criteria from the available user history, making the decision basis inferable and relatively stable even with sparse interactions. 

\paragraph{\textsc{P-Check} consistently enhances reward accuracy across diverse LLM-Judges.}
To validate the transferability of our framework, we apply \textsc{P-Check} across diverse LLM-Judges, ranging from open-weight models (Qwen) to proprietary models (GPT). From the results in Table ~\ref{tab:pcheck_backbones}, \textsc{P-Check} consistently improves reward accuracy across all judges, indicating that even frontier models benefit significantly from explicit checklists that steer their knowledge toward user-specific preferences.
In particular, a compact 3B checklist generator within \textsc{P-Check} boosts the performance of much larger judges, highlighting that guidance for \textit{what to evaluate} is just as critical as the judge's intrinsic capacity for the personalized reward prediction.
% preamble에 추가 필요:
% \usepackage[table]{xcolor}
% \usepackage{colortbl}

\begin{table}[h]
\centering
\small
\setlength{\tabcolsep}{2.5pt}
\renewcommand{\arraystretch}{1.08}
\begin{tabularx}{\columnwidth}{lccc}
\toprule
& \textbf{PRISM-P.} & \textbf{ARENA-P.} & \textbf{BESPOKE-M.} \\
\midrule
Qwen3-8b & \pmv{55.14}{2.16} & \pmv{57.41}{2.51} & \pmv{59.23}{3.05} \\
\rowcolor{black!7}\textbf{+ \textsc{P-Check}} & \pmv{\textbf{63.71}}{1.95} & \pmv{\textbf{59.98}}{2.17} & \pmv{\textbf{70.36}}{2.87} \\
\midrule
Qwen3-13b & \pmv{59.76}{2.37} & \pmv{58.89}{2.59} & \pmv{64.14}{3.40} \\
\rowcolor{black!7}\textbf{+ \textsc{P-Check}} & \pmv{\textbf{63.23}}{2.03} & \pmv{\textbf{63.62}}{2.20} & \pmv{\textbf{79.16}}{3.15} \\
\midrule
GPT-4o-mini & \pmv{56.07}{1.98} & \pmv{59.86}{2.58} & \pmv{60.23}{3.05} \\
\rowcolor{black!7}\textbf{+ \textsc{P-Check}} & \pmv{\textbf{63.40}}{2.03} & \pmv{\textbf{62.31}}{2.35} & \pmv{\textbf{76.51}}{2.59} \\
\midrule
GPT-4o & \pmv{58.94}{2.34} & \pmv{60.06}{2.09} & \pmv{63.27}{2.86} \\
\rowcolor{black!7}\textbf{+ \textsc{P-Check}} & \pmv{\textbf{64.83}}{2.10} & \pmv{\textbf{69.36}}{2.30} & \pmv{\textbf{77.66}}{3.19} \\
\bottomrule
\end{tabularx}
\vspace{-0.25cm}
\caption{Performance comparison of diverse LLM-Judges on reward prediction with and w/o \textsc{P-Check}.}
\label{tab:pcheck_backbones}
\vspace{-0.6cm}
\end{table}

\paragraph{Reward output of \textsc{P-Check} boosts personalized alignment of the policy model.}
Beyond the accuracy of reward prediction, we examine the utility of \textsc{P-Check} in steering policy models toward better personalized alignment.
We use BESPOKE to evaluate personalized generation under two popular alignment strategies: \textit{Best-of-$N$} (\textit{BoN}) and \textit{DPO}.
Specifically, we measure generation quality against human-annotated gold references using ROUGE, METEOR, and BESPOKE-EVAL.
For \textit{BoN}, we sample 10 roll-outs per query from the policy (\textit{i.e.}, Llama-3-8b, GPT-4o-mini) and select the best response using each reward model.
For \textit{DPO}, since BESPOKE only provides a test split, we first construct a synthetic query and pair-wise policy roll-outs, then label preferences via each reward model to train a Llama3-8B policy.
We provide detailed setups in Appendix~\ref{app:align_details}.
From the results in Table~\ref{table:align}, we observe that incorporating \textsc{P-Check} as reward model for policy optimization yields the best generation quality across all settings. 
Notably, these improvements are consistent across both inference-time scaling (BoN) and parameter optimization (DPO), which demonstrates the versatility of \textsc{P-Check} as a robust reward signal.

\begin{figure}[ht]
\centering
\includegraphics[width=\linewidth]{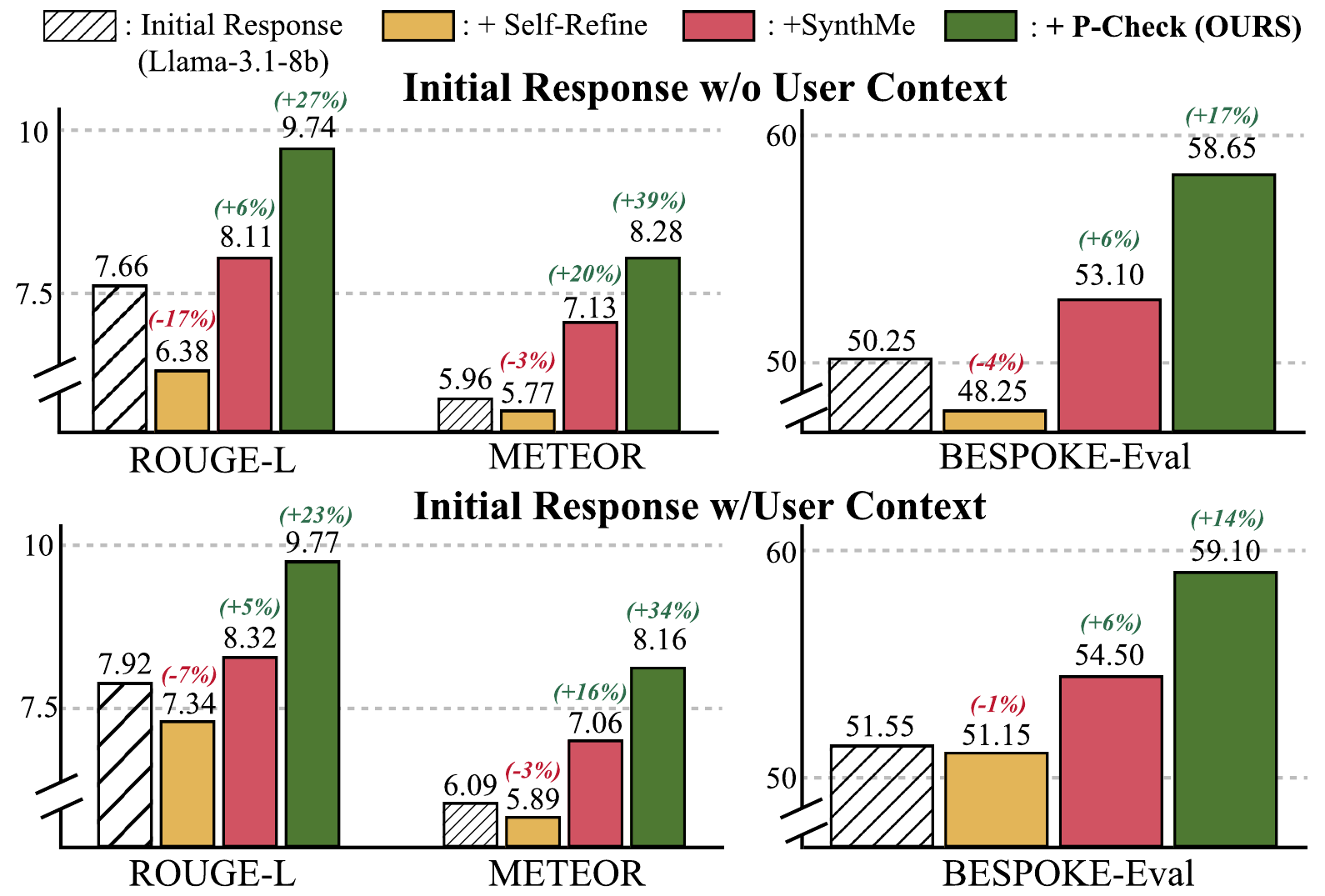}
\vspace{-0.5cm}
\caption{Results on refining the output of policy model with feedback from \textsc{P-Check}'s inferred checklist.}
\label{fig:feedback}
\vspace{-0.5cm}
\end{figure}

\paragraph{Checklist of \textsc{P-Check} provides useful feedback for the policy model.} We further test whether the personalized checklist inferred by \textsc{P-Check} can act as useful verbal feedback for improving a policy model’s personalized generation. 
Specifically, on BESPOKE we first generate an initial response from the Llama-3.1-8B policy either with and without user context (interaction history), then prompt the policy to refine the initial response along with the inferred checklist. As baselines, we compare \textsc{P-Check} against Self-Refine~\citep{madaan2023selfrefine} and SynthMe (Persona-based).
As shown in Figure~\ref{fig:feedback}, \textsc{P-Check} delivers the largest improvements across both settings, while Self-Refine often yields negative changes and SynthMe provides smaller gains.
These results suggest that the checklist of \textsc{P-Check} provides actionable guidance for personalized refinement that the policy can directly apply, which enables lightweight personalization without additional policy parameter updates.

\paragraph{Ablation Study.} 
% To understand the impact of our core components, we conduct an ablation study by removing the two main steps of our Preference-Contrastive Criterion Weighting.
To understand the impact of our core components, we ablate the two main steps of our Preference-Contrastive Criterion Weighting.
As reported in Table~\ref{tab:ablation}, ablating either component leads to a consistent performance drop, with the removal of Saliency Scoring causing the most significant degradation. 
This shows that capturing unique user preferences via contrastive sampling and identifying key criteria via saliency scoring are both essential for the overall efficacy of the model.
% preamble: \usepackage{booktabs,tabularx}

\begin{table}[h]
\centering
\small
\setlength{\tabcolsep}{0.9pt}
\renewcommand{\arraystretch}{1.1}
\begin{tabularx}{\columnwidth}{lccc}
\toprule
& \textbf{PRISM-P.} & \textbf{ARENA-P.} & \textbf{BESPOKE-M.} \\
\midrule
\textbf{\textsc{P-Check} (full)} & \pmv{\textbf{65.11}}{2.44} & \pmv{\textbf{61.56}}{1.85} & \pmv{\textbf{75.48}}{2.27} \\
(-) Intr-Usr Smpl.     & \pmv{63.46}{2.56} & \pmv{59.56}{1.90} & \pmv{72.23}{2.35} \\
(-) Saliency Scr.        & \pmv{59.98}{1.90} & \pmv{58.46}{2.16} & \pmv{66.68}{2.35} \\
\bottomrule
\end{tabularx}
\vspace{-0.2cm}
\caption{Ablation results of \textsc{P-Check}.}
\label{tab:ablation}
\vspace{-0.6cm}
\end{table}

\section{Related Works}
\label{sec:relatedwork}
% We provide broader related works in Appendix~\ref{app:relwork}.

\paragraph{Personalized Reward Modeling.}
Early approaches typically operate under constrained settings with predefined preference dimensions, estimating user-specific mixtures over fixed axes~\citep{rame2023rewarded,jang2023personalizedsoupspersonalizedlarge,chen2025pad}. 
More recent works learn unconstrained preferences directly from interaction histories: \citet{zhao2024group} predicts group-level preferences from embeddings of prior preference judgments, \citet{poddar2024personalizing} explicitly learns latent user embeddings inferred from interaction history, and \citet{chen2025pal} learns pluralistic preference prototypes with user-specific mixing weights in a shared latent space. 
Alternatively, non-parametric approaches like \citet{ryan-etal-2025-synthesizeme} leverage LLMs to synthesize natural language personas from interaction history and optimize the prompt of judge model conditioned on these personas for reward scoring.
In contrast, \textsc{P-Check} shifts the modeling target from scalar outcomes to the evaluation logic itself, generating a query-specific checklist that provides dynamic guidance for reward prediction. 
% By directly learning the supporting evidence for each judgment, P-Check provides more transparent rationales for model decisions and improves generalization without relying on spurious correlations ~\citep{d}.

% \paragraph{Checklist-guided Evaluation.}
\paragraph{Checklist-guided Evaluation.}
In parallel, evaluation paradigms have shifted towards fine-grained assessment to enhance reliability and interpretability. 
Benchmarks such as \citet{ye2024flask} and evaluators like \citet{kim2024prometheus} decompose response quality into detailed criteria, scoring outputs along multiple axes.
Recent works extend this utility to reward signals; frameworks like \citet{viswanathan2025checklists} and \citet{gunjal2025rubricsrewardsreinforcementlearning} demonstrate that aggregating checklists into reinforcement learning yields more stable supervision. 
However, these methods remain largely user-agnostic: criteria are typically predefined for a task or generated solely based on the instruction, disregarding individual user histories. 
\textsc{P-Check} bridges this gap by generating history-aware checklists with personalized saliency, distinguishing essential constraints from optional preferences to provide a fine-grained reward signal tailored for the personalized alignment.

\paragraph{Personalization of Large Language Models.}
General approaches on LLM personalization spans a broad spectrum of techniques aimed at adapting model behavior to individual users.
Retrieval-based methods ~\citep{salemi-etal-2024-lamp, kim-yang-2025-shot} retrieve user-specific history to augment the input context, while modular approaches like \citet{liu-etal-2025-llms} trains dynamic memory plugs to adapt the model.
To internalize preferences, recent works focus on deeper user modeling: \citet{balepur-etal-2025-whose} enhance preference tuning by leveraging inferred user personas, and \citet{zhao2025nextquillcausalpreferencemodeling} employs causal modeling to capture the underlying determinants of user preferences.
At the inference stage, decoding-time strategies ~\citep{chen2025pad, kim-etal-2025-drift} steer generation by contrasting personalized logits against generic ones.
Furthermore, interactive frameworks such as \citet{wu-etal-2025-aligning} and \citet{zhao2025teaching} user simulation~\citep{kim2025stopplayingguessinggame} to evolve the alignment policy alongside the user.
Despite the remarkable efficacy, these approaches typically entangle preference modeling with response generation, expecting the model to implicitly derive constraints from noisy history while decoding.
In contrast, \textsc{P-Check} decouples the formulation of user-specific evaluation criteria from the conditioned response generation process.
By transforming raw interaction logs into actionable specifications, our approach enables personalization grounded in transparent and verifiable criteria rather than implicit patterns.

\paragraph{Generative Reward Modeling.}
Moving beyond conventional reward models that output a single scalar score, Generative Reward Models leverage the reasoning capabilities of LLMs to output explanations alongside evaluations.
\citet{zhang2024generative} establish this paradigm by formulating verification as a next-token prediction task, demonstrating remarkable generalizability over standard regression. 
Recent advancements extend this foundation: \citet{xu2025genarm} employs autoregressive reward modeling for test-time alignment, and \citet{ye2025learning} optimizes judge models to better approximate human preference distributions. 
To further enhance granularity and scalability, \citet{liu2025inferencetimescalinggeneralistreward} introduces critique tuning to scale high-quality principles, while some works analyze reasoning trajectories to verify process-level, intermediate steps~\citep{chae2025webshepherd, zhao2025genprmscalingtesttimecompute, xiong2025stepwiserstepwisegenerativejudges}.
However, these works primarily target objectively verifiable domains such as mathematics or code, where evaluation relies on universal correctness.
\textsc{P-Check} extends this generative verification to the subjective problem of personalized alignment.
Instead of generating a generic chain-of-thought for correctness, \textsc{P-Check} generates user-conditional checklists, effectively serving as a personalized verifier that reasons about alignment based on individual interaction history.

\section{Conclusion}
\label{sec:conclusion}
We propose \textsc{P-Check}, a novel personalized reward modeling framework designed to train a plug-and-play checklist generator that synthesizes dynamic evaluation criteria for guiding the personalized reward prediction. 
Experimental results show that \textsc{P-Check} not only improves reward accuracy but also enhances downstream personalized generation. 
We believe \textsc{P-Check} establishes a robust foundation for developing more interpretable and reliable reward model aligned with diverse user needs.

\section*{Limitations}
\label{sec:limitation}
While \textsc{P-Check} achieves strong gains compared to existing approaches, it also introduces several limitations that point to promising future directions.
First, \textsc{P-Check} assumes that a user’s decision basis can be represented as a set of discrete, natural-language criteria. However, some preference factors are hard to externalize as explicit rules (e.g., subtle tone, style, pacing, or ``feel'') and may be only partially captured by a checklist interface. In such cases, explicit criteria may underrepresent fine-grained preference signals that are better modeled implicitly. A natural extension is to explore how implicit preference factors can be modeled as evaluation criteria, \emph{i.e.}, in a form that a judge can apply reliably during reward prediction.

Second, since reward prediction is steered by the synthesized checklist, any inaccuracy in the generated criteria can directly affect the final judgment. In particular, since the checklist serves as the evaluation interface, any flaws in it can systematically distort the criterion-wise scoring and propagate to the final reward. While we apply rejection sampling to filter out low-quality checklist in training data, and our saliency weighting helps reduce the influence of weakly discriminative items, this does not guarantee that every generated criterion is faithful to the query and the user evidence. Ensuring criterion validity and preventing such criteria from propagating into reward prediction remains an important open challenge.

Lastly, our training pipeline involves multiple steps (e.g., summarizing $GP_u$, constructing contrastive sets, and computing saliency-based supervision), which introduces non-trivial training-time cost and system complexity. However, these components can be largely minimized through offline user management: user-level summaries and inter-user contrast computations can be precomputed and cached, and saliency labeling can be performed as a one-time preprocessing step for each checklist. Moreover, in our experiments, test-time latency remains relatively modest (Table~\ref{tab:inference_time_acc}) since inference only requires generating a compact checklist and scoring a small set of criteria. Nevertheless, improving the overall efficiency of the training pipeline remains an important practical direction.

\section*{Ethical Consideration}
\label{sec:ethical}
LLM-based generation can produce incorrect or hallucinated outpus~\citep{seo-etal-2024-unveiling} and may contain harmful, biased, or offensive language~\citep{kim-etal-2024-verifiner,seo-etal-2025-mt}. This is particularly relevant to \textsc{P-Check} because the framework relies on generated artifacts (e.g., $GP_u$, checklists, and synthetic contrastive responses) as intermediate signals for reward prediction, and any problematic content in these artifacts could propagate to downstream judgments or policy optimization. However, we believe this risk is largely minimized in our study through controlled use of established public benchmarks and quality controls during synthetic artifact construction.

\textbf{Data sources, licensing, and privacy.}
Our experiments are conducted on publicly available personalized reward benchmarks, including PRISM, ChatbotArena, and BESPOKE.
We use these datasets under their respective licenses and intended research use.
Since personalization benchmarks are derived from user interaction histories, privacy is a central concern.
We do not attempt to de-anonymize users, and we treat user histories and derived summaries ($GP_u$) as sensitive signals.
Accordingly, we avoid releasing any additional user-identifying information beyond what is already included in the original benchmark distributions, and we ensure that \textsc{P-Check} is evaluated in a controlled research setting rather than deployed on real user data collected by us.

\textbf{Mitigating harmful or low-quality generations and human safeguards.}
We apply rejection sampling and quality filtering when constructing $GP_u$ and checklist training data to discard low-quality or malformed generations, and we leverage the checklist interface to make reward computation more inspectable. 
For preliminary analyses that require human verification of generated checklists, graduate-student annotators follow written guidelines and we limit the daily workload to reduce fatigue-related artifacts.
All human-facing annoation is restricted to the scope of verifying checklist validity for research purposes, and does not involve collecting new personal data.

\textbf{Broader impacts and safe personalization.}
Finally, personalized reward modeling raises the broader risk that a system may over-optimize for user-specific preferences that could be harmful, discriminatory, or otherwise unsafe.
\textsc{P-Check} should therefore be viewed as a component for improving preference fidelity and transparency of reward prediction, not as a substitute for safety constraints.
In practical deployments, it is necessary to combine personalization mechanisms with safety policies.

\section*{Acknowledgement}
This work was supported by the IITP grants funded by the Korea government (MSIT) (No.RS-2020-II201361; RS-2024-00457882, AI Research Hub Project; RS-2026-25520654), and computational resources from AWS Trainium via the Theta EdgeCloud platform.

\bibliography{custom}

\newpage
\appendix

\clearpage
\section{Appendix}

\subsection{Experimental Details}

\subsubsection{Preliminary Analysis}
\label{app:prelim_details}
To conduct the preliminary analyses described in Section~\ref{sec:prelim}, we sample 100 users from the PRISM-Personalized dataset in PersonalRewardBench~\citep{ryan-etal-2025-synthesizeme}.
Specifically, we retain only those instances that provide at least four non-overlapping chosen–rejected response pairs for the same user and query.
This constraint is necessary to conduct the experiment in Analysis 2 without information leakage, ensuring that the pairs used to derive the evaluation criteria are disjoint from the pairs used to test the model's preference selection.
For all experiments, consistent with the main experimental setup, we report the mean performance over 5 runs.
Additionally, we visualize the 95\% confidence intervals in the corresponding figures to demonstrate statistical reliability. 

To generate the Oracle Checklist, we prompt GPT-4o with the user's interaction history, the current query, and the ground-truth chosen response ($y^+$), instructing it to construct evaluation criteria that can justify why $y^+$ aligns with the user's preference based on the provided history.
Conversely, for the Counter-Preference Checklist used in Analysis 1, we prompt GPT-4o to justify the rejected response ($y^-$).
As a baseline for Analysis 2, we also generate an Oracle Persona. Unlike the query-specific checklist, the persona is derived solely from the user's history and labeled pairs to create a static descriptive profile.

To ensure the validity of the generated checklist and persona, we perform manual human verification for all generated checklists and personas.
We verify each sample on two dimensions: (1) whether the criteria are plausibly grounded in the user's interaction history, and (2) whether they logically justify the respecive target response.
Samples failing these verificatoon process are manually corrected or regenerated.
For the preference selection experiment in Analysis 2, we strictly split a held-out set to prevent data leakage.
Specifically, for a given user and query, we use one set of chosen–rejected pairs to generate the Oracle Checklist (or Persona) and evaluate the model's performance on a completely disjoint set of pairs.
This ensures that the reported improvements reflect the model's ability to apply the inferred criteria to new candidates, rather than memorization of the source instances.

\subsubsection{Implementation Details of \textsc{P-Check}}
\label{app:implement_detail}
\paragraph{Collecting Checklist Training Data.}
In the training data collection phase, we employ rejection sampling~\citep{yuan2023scalingrelationshiplearningmathematical} to filter out low-quality generations and secure reliable data for training.
Specifically, we provide the generated $GP_u$ and the raw checklist $C_{u,q}$ as additional input context to the LLM-judge (Llama-3.1-8b) and perform zero-shot inference.
Subsequently, we only retain the samples where the judge successfully assigns a higher reward score to the chosen response compared to the rejected one.
For instances that fail this verification, we repeat the generation process to ensure the construction of a high-quality training dataset.

\begin{figure}[ht]
\centering
\includegraphics[width=\linewidth]{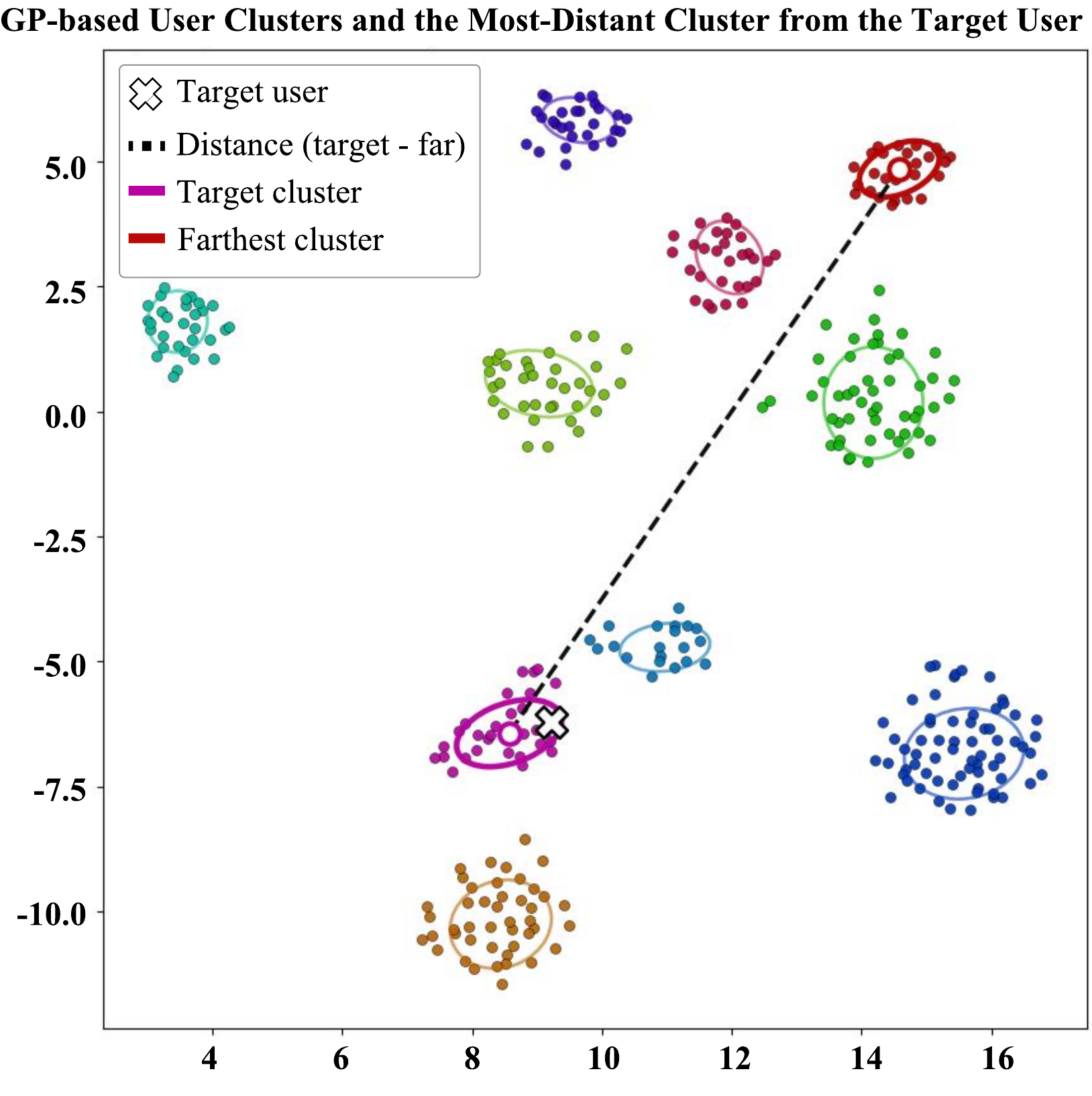}
\vspace{-0.5cm}
\caption{Visualization of the coarse-grained filtering step. We cluster users based on their static $GP$ embeddings using K-Means Clustering. For a target user (marked with `X'), we identify the candidate pool for contrastive sampling by selecting the cluster whose centroid is farthest from the target user's cluster centroid.}
\label{fig:cluster}
\vspace{-0.5cm}
\end{figure}
\begin{figure}[h]
\centering
\includegraphics[width=\linewidth]{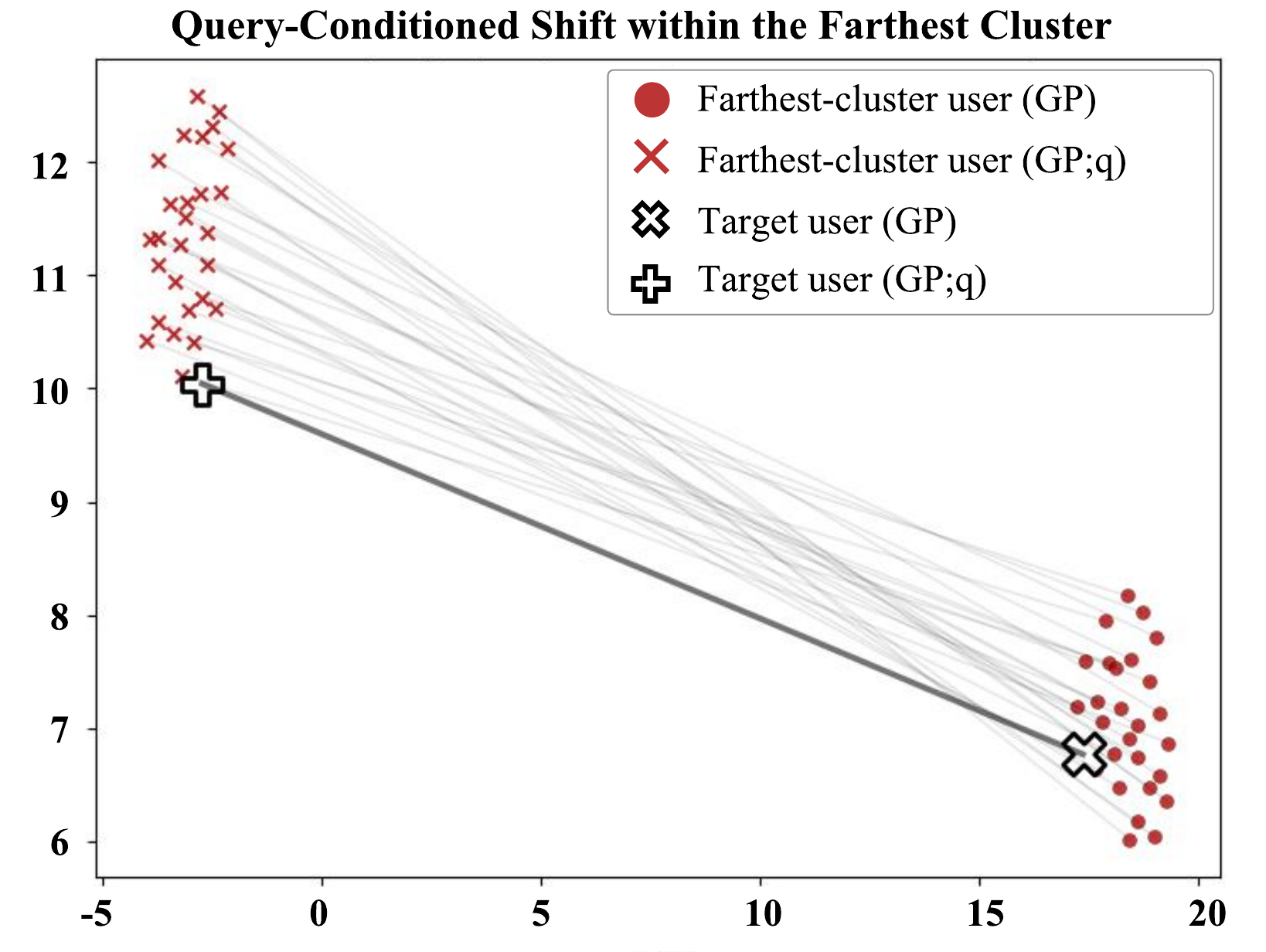}
\vspace{-0.6cm}
\caption{Visualization of query-conditioned embedding shift in the farthest cluster. Lines connect static preferences ($GP$, dots) to query-conditioned representations ($\text{Enc}(GP, q)$, crosses) in a shared 2D space.}
\label{fig:queryshift}
\vspace{-0.6cm}
\end{figure}
\paragraph{Inter-User Contrastive Sampling}
For the inter-user contrastive sampling, we utilize Qwen-3-embedding 0.6B as the embedding model.
This choice reflects the model's balance of efficiency and performance, which is essential for managing a vast user pool in practical deployment scenarios.
We select the number of clusters (10 groups) based on the Silhouette Coefficient optimization performed on the PRISM training users.
From the identified distant cluster $\mathcal{V}$, we select the top $K=3$ most distant users to augment the negative set.
For the generation of $y^-_v$, we employ two different LLMs (Qwen-3-13B, GPT-4o) to prevent bias toward the specific style of a single model, and enable the checklist generator to robustly learn priorities across diverse response patterns.
We present visualization examples on the PRISM dataset in Figure~\ref{fig:cluster} and Figure~\ref{fig:queryshift}.

\begin{figure}[h]
\centering
\includegraphics[width=\linewidth]{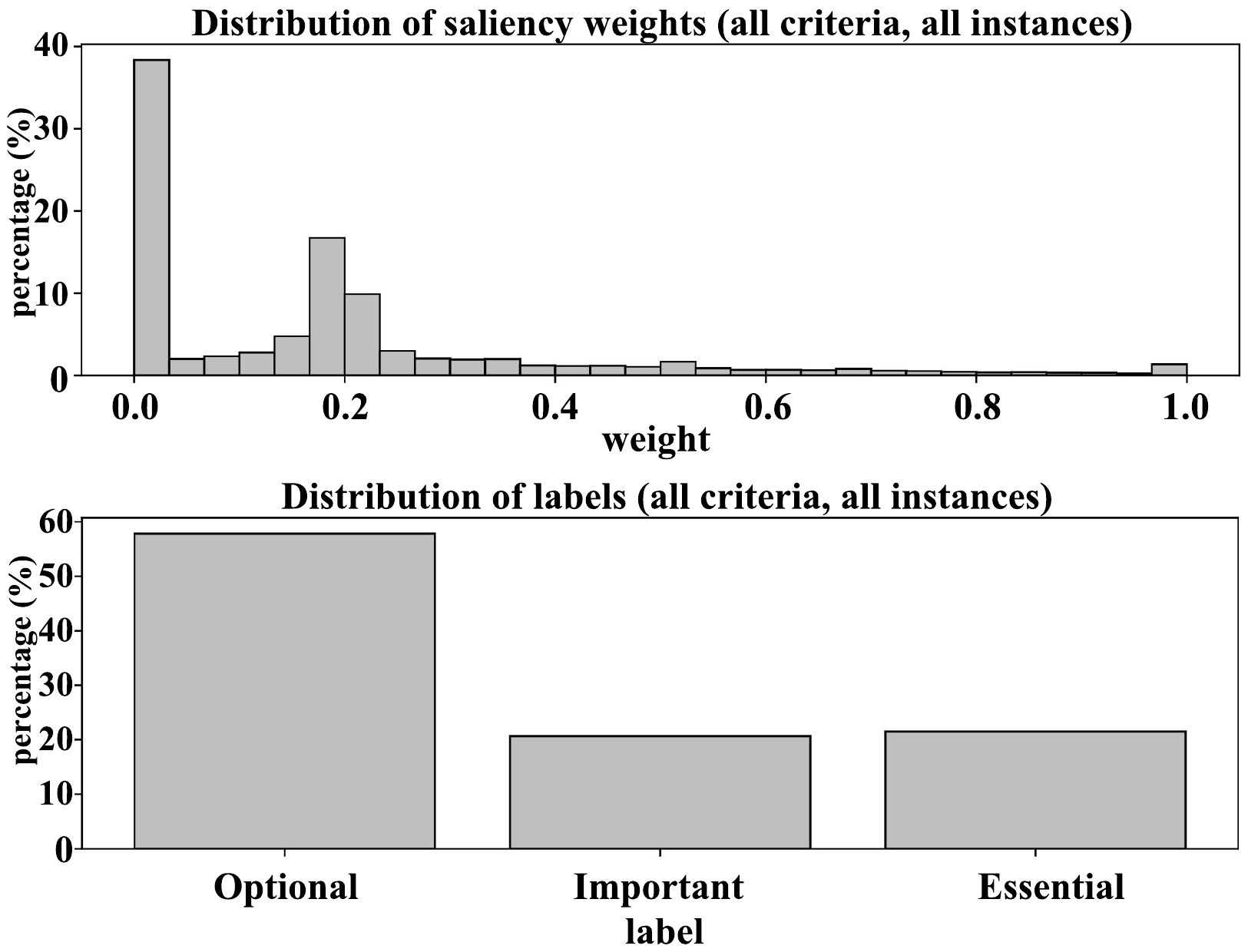}
\vspace{-0.3cm}
\caption{Distribution of saliency weights (Upper) and the corresponding categorical labels (Lower).}
\label{fig:label_dist}
\vspace{-0.3cm}
\end{figure}
\paragraph{Personalized Saliency Scoring}
For the personalized saliency scoring, we employ Llama-3.1-8B as the scoring model.
We evaluate the adherence of a response $y$ to each checklist criterion $c_k$ on a scale of 1 to 10, with a temperature of 1.0.
For the calculated saliency scores, we convert them into categorical labels—\texttt{Essential}, \texttt{Important}, and \texttt{Optional}, adopting a strategy similar to \citet{gunjal2025rubricsrewardsreinforcementlearning}.
Specifically, we sort the criteria in $C_{u,q}$ by their weights in descending order and traverse the list while tracking the cumulative sum.
We set the cumulative probability thresholds $\tau_1$ and $\tau_2$ to 0.4 and 0.9, respectively. We provide an ablation study of 
$\tau_1$ and $\tau_2$ in Table~\ref{tab:tau_sensitivity}, reporting both the label distribution over O/I/E and the final accuracy when training \textsc{P-Check} under each hyperparameter setting.
The distribution of the assigned labels is presented in Figure~\ref{fig:label_dist}.

\paragraph{Training Checklist Generator $\phi$}
For the training of the checklist generator $\phi$, we employ Llama-3.2-3B-Instruct as the backbone model.
The training is conducted on 8 NVIDIA A6000 GPUs for 3 epochs.
We utilize a per-device batch size of 2 with 16 gradient accumulation steps.
The model is optimized using a learning rate of $2 \times 10^{-4}$.

\paragraph{Inference with the Generated Checklist}
At inference time, we first use the trained generator to synthesize the checklist, and subsequently compute the reward using the LLM-judge conditioned on this generated checklist.
We use a temperature of 1.0 for all generation steps.
To aggregate the scores, we map the generated categorical weight labels (\texttt{Essential}, \texttt{Important}, \texttt{Optional}) into numerical scalars based on validation accuracy.
We provide an ablation study on these weight assignments in Table~\ref{tab:weight_ablation}.
From the results, we identify that weights of 1.0, 0.7, and 0.3 for \texttt{Essential}, \texttt{Important}, and \texttt{Optional} respectively yield the optimal performance.
However, we also observe that the final reward accuracy is not significantly sensitive to these specific hyperparameter values, suggesting the robustness of our framework.
\begin{table}[h]
\centering
\small
\begin{tabular*}{\columnwidth}{@{\extracolsep{\fill}}cccccc@{}}
\toprule
$\tau_1$ & $\tau_2$ & O & I & E & Acc. \\
\midrule
0.6 & 0.9 & 70.92 & 7.55  & 21.52 & 63.27 \\
\textbf{0.4} & \textbf{0.9} & 57.81 & 20.67 & 21.52 & 65.11 \\
0.2 & 0.9 & 52.58 & 25.96 & 21.52 & 65.89 \\
0.4 & 0.8 & 57.81 & 14.58 & 27.59 & 64.42 \\
0.4 & 0.7 & 57.81 & 13.06 & 29.12 & 65.04 \\
0.4 & 0.6 & 57.81 & 8.77  & 33.05 & 64.09 \\
\bottomrule
\end{tabular*}
\caption{Sensitivity analysis over saliency thresholds $\tau_1$ and $\tau_2$. We report the induced label distribution over Optional (O), Important (I), and Essential (E), along with final reward prediction accuracy.}
\label{tab:tau_sensitivity}
\end{table}
\begin{table}[t]
    \centering
    \small
    \setlength{\tabcolsep}{6pt}
    \renewcommand{\arraystretch}{1.0}

    \begin{adjustbox}{max width=\columnwidth,center}
    \begin{tabular}{@{}ccccc@{}}
    \toprule
    \textbf{Essential} & \textbf{Important} & \textbf{Optional} & \textbf{Val Acc.} & \textbf{Test Acc.} \\
    \midrule
    1.0 & 0.9 & 0.7 & 61.24 & 62.85 \\
    1.0 & 0.9 & 0.5 & 61.89 & 61.10 \\
    1.0 & 0.9 & 0.3 & 62.15 & 62.45 \\
    \midrule
    1.0 & 0.8 & 0.6 & 62.03 & 63.22 \\
    1.0 & 0.8 & 0.4 & 63.58 & 62.98 \\
    1.0 & 0.8 & 0.2 & 62.40 & 63.75 \\
    \midrule
    1.0 & 0.7 & 0.5 & 64.10 & 64.50 \\
    \textbf{1.0} & \textbf{0.7} & \textbf{0.3} & \textbf{64.27} & \textbf{65.11} \\
    1.0 & 0.7 & 0.1 & 63.25 & 65.20 \\
    \midrule
    1.0 & 0.6 & 0.4 & 63.80 & 63.05 \\
    1.0 & 0.6 & 0.2 & 62.45 & 62.60 \\
    \midrule
    1.0 & 0.5 & 0.3 & 63.92 & 61.88 \\
    \bottomrule
    \end{tabular}
    \end{adjustbox}

    \caption{Hyperparameter grid search for criterion weight assignments, conducted on the PRISM dataset using the Llama-3.1-8B-Instruct as the judge model. We fix the weight of \texttt{Essential} to 1.0 and vary \texttt{Important} and \texttt{Optional}. The configuration selected for our final model is marked in bold.}
    \label{tab:weight_ablation}
\end{table}

\subsubsection{Datasets}
\label{app:dataset}

\paragraph{PRISM-Personalized.}
We utilize the PRISM-Personalized dataset from PersonalRewardBench~\citep{ryan-etal-2025-synthesizeme}, which serves as our primary in-distribution benchmark with data from 723 users.
Originally derived from PRISM~\citep{kirk2024the}, this dataset maps detailed survey responses onto multi-turn conversations emphasizing values and controversial topics.
While the original PRISM collects $N$-way comparisons with cardinal ratings (1–100), PersonalRewardBench converts these into pairwise formats, filtering out pairs with less than a 10\% quality difference.
To benchmark personalized reward modeling accuracy, this dataset is constructed by a rigorous filtering pipeline, which retains users with sufficient interaction history and selects for queries with high personalization potential and low consensus among general LLM judges, thereby isolating subjective or controversial instances.

\paragraph{ChatbotArena-Personalized.}
We also employ the ChatbotArena-Personalized dataset from PersonalRewardBench~\citep{ryan-etal-2025-synthesizeme} as an out-of-distribution evaluation source, comprising data from 131 users.
These are originally sourced from Chatbot Arena~\citep{chiang2024chatbotarenaopenplatform}, which facilitates in-the-wild, open-ended conversations where users blindly compare two anonymous LLMs.
Similar to PRISM-Personalized, this subset is collected through the same data filtering pipeline, which limits inclusion to users with sufficient history and selects for examples with high personalization potential and high disagreement.

\paragraph{BESPOKE-MetaEval.}
We use BESPOKE-MetaEval~\citep{kim2025bespokebenchmarksearchaugmentedlarge}, a more recent and challenging benchmark that provides both user interaction history and search history. 
For the reward accuracy experiments, we utilize the Meta-Evaluation set provided by the benchmark. 
In this set, each user query is associated with multiple response candidates, each annotated with 1--5 scale scores across various personalization aspects. 
To construct pairwise preferences, we calculate the average score for each response across these aspects and sort them to form chosen--rejected pairs. 
Since baseline methods such as GPO, PAL, VPL, and SynthMe require access to preference labels even at test time (as few-shot examples or for optimization), we split the meta-evaluation set for each user, allocating 20\% of the preference pairs as a supporting set and retaining the remaining data as a held-out test set.

\subsubsection{Baselines}
\label{app:baselines}

\paragraph{Default.}
In the Default setting, we directly prompt the LLM-judge to identify the response that better aligns with the user's preference.
The model conditions its decision solely on the provided user interaction history without utilizing any additional retrieval mechanisms or intermediate reasoning.

\paragraph{Memory.}
For Memory, we adopt a retrieval-augmented approach following prior works in personalized generation~\citep{salemi-etal-2024-lamp, ryan-etal-2025-synthesizeme}.
We utilize Qwen3-Embedding-0.6B to retrieve the top-5 most similar interaction history instances (consisting of the query, chosen response, and rejected response) based on the current query.
These retrieved instances serve as in-context demonstrations to guide the judge model in predicting the user's preference.

\paragraph{SynthesizeMe.}
We follow the official implementation of SynthesizeMe~\citep{ryan-etal-2025-synthesizeme}, which optimizes a user persona and selects user-level demonstrations to steer the model.
For experiments on the BESPOKE-MetaEval, we adapt this baseline to ensure a fair comparison, since the official setting of BESPOKE only provides implicit feedback in user history without pair-wise labels.
Specifically, we employ the same generated general preference $GP_u$ used in \textsc{P-Check} as the user persona and employ all the supporting pairs, excluding the held-out test set, as in-context demonstrations for each test user to guide the preference selection.

\paragraph{CoT-\textit{distill}.}
CoT-distill represents a chain-of-thought distillation approach~\citep{hsieh-etal-2023-distilling} where the model learns to generate free-form rationales.
Unlike \textsc{P-Check}, which structures context as a checklist, this baseline trains a reasoner to output a step-by-step natural language justification for the chosen response.
We use the same training data configuration ($GP_u$, query, and preference pairs) and backbone model (Llama-3.1-8B-Instruct) as our checklist generator.
We also apply rejection sampling to ensure the quality of the training rationales.
At inference time, the trained reasoner synthesizes a rationale based on $GP_u$ and the query, which then serves as additional context for the judge to select the better personalized response.

\paragraph{GPO.}
GPO (Group Preference Optimization)~\citep{zhao2024group} adapts LLM outputs to specific group preferences using a meta-learning framework.
We follow the implementation of \citet{ryan-etal-2025-synthesizeme} to handle pairwise preferences by embedding a prompt containing the user context and candidate responses with Llama3-8b-Instruct and Llama3-3b-Instruct.
A separate Transformer-based preference module then processes these embeddings to predict a binary preference label.
We optimize the module using Adam with a learning rate of $3 \times 10^{-5}$ and cosine annealing for 200,000 steps, using mean-pooling for embeddings.

\paragraph{VPL.}
VPL (Variational Preference Learning)~\citep{poddar2024personalizing} treats preference modeling as a latent variable problem to address the limitation of assuming a single utility function.
It estimates a hidden variable $z$ representing user context via variational inference from pairwise annotations.
The reward model conditions on this latent space to capture multi-modal preferences.
We extend the original implementation to support general preference learning, enabling the model to refine the latent variable $z$ at test time for personalization.
We use Llama-3-3B-Instruct and Llama-3-8B-Instruct models as base encoders and train the model using a composite objective of log-sigmoid loss and KL divergence regularization.
Optimization is performed using AdamW with a learning rate of $3 \times 10^{-4}$.

\paragraph{PAL.}
PAL (Pluralistic Alignment Framework)~\citep{chen2025pal} models diverse user values by representing each user as a mixture of "prototypical preference points" within a transformed representation space.
It jointly learns a mapping function and prototypes to infer mixture weights, allowing for efficient personalization.
We follow the original paper and implement the PAL-B variant with frozen LLM parameters, setting the dimension of preference embeddings equal to the hidden dimension of the LLM encoder.
For the projection architecture, we utilize a 2-layer MLP with GELU activations and Gaussian initialization, while disabling the learnable temperature.
We set the number of prototypical points $K$ to 2 and conduct training with a batch size of 1.

\paragraph{BT + SynthesizeMe.}
BT + SynthesizeMe combines a standard Bradley-Terry reward model with the persona-based context from SynthesizeMe.
We fine-tune the scalar reward model using a LoRA adapter with a rank of 32 and the standard contrastive reward modeling loss provided by the HuggingFace TRL library.
We set the per-device batch size to 1, train for 2 epochs with a learning rate of $1 \times 10^{-5}$, and use a maximum sequence length of 8192 tokens.
We incorporate the generated persona and demonstrations into the input prompts during both the fine-tuning phase and the inference stage to condition the reward prediction on the synthesized user context.

\subsubsection{Experimental Details on Personalized Alignment}
\label{app:align_details}

For the personalized generation experiments, we evaluate performance by comparing the model-generated responses against the human-annotated gold references (gold information need) provided in the BESPOKE benchmark.
We employ two lexical similarity metrics, ROUGE-L and METEOR, alongside the official BESPOKE-EVAL metric (denoted as B.Eval in Table~\ref{table:align}).
Specifically, we measure four personalization aspects (\textit{i.e.,} Need Alignment, Content Depth, Tone, and Style) and report the average score across these dimensions to assess comprehensive alignment quality.

In the default setting, the policy model receives the general preference summary derived from the user history and generates parallel roll-outs for each query.
We then align these outputs using various baseline reward models and \textsc{P-Check}.
For Best-of-$N$ selection, we generate $N=10$ candidate responses per test query using a temperature of 1.0.
We then select the final prediction as the response assigned the highest score by each reward model.
For baselines that operate as pairwise judges or selectors, such as CoT-distill and SynthMe, we prompt them to identify the most personalized response from the entire candidate pool directly.

Since BESPOKE only provides a test split, we adopt a synthetic data generation pipeline for the DPO experiments, following the setting in recent works~\citep{yu2025rip, xu2025magpie}.
To construct the alignment training data, we first synthesize queries by prompting GPT-4o-mini, using queries from the splitted supporting set (excluding the held-out test set) as few-shot demonstrations.
For each synthesized query, we sample pairwise responses from the policy model (Llama-3.1-8B).
We then label these pairs using each reward model, assigning the response with the higher predicted reward as chosen and the other as rejected to form the preference training set for DPO.

\subsection{Further Analysis}

\subsubsection{Cost Analysis}
To evaluate the practical efficiency of \textsc{P-Check}, we measure the total end-to-end wall-clock time for inference on the PRISM test set. 
To ensure a fair and consistent comparison, all experiments are conducted using the vLLM ~\citep{kwon2023efficientmemorymanagementlarge} on a cluster of 8 NVIDIA A6000 GPUs. 
As summarized in Table~\ref{tab:inference_time_acc}, \textsc{P-Check} introduces a moderate test-time overhead of approximately 9 minutes over the Llama-3-8B baseline, yet this is offset by a substantial accuracy gain of +12.31 points.
Notably, when compared to the larger Qwen3-13B model, \textsc{P-Check} achieves a higher accuracy (+5.35 points) while maintaining a comparable inference time, despite a slight increase of 1 minute and 55 seconds.
This indicates that our approach provides a better performance-to-cost trade-off than simply scaling up the model parameters.
Furthermore, \textsc{P-Check} is both faster (saving ~8 minutes) and more accurate (+2.37 points) than BT+SynthMe, showing that our structured checklist generation provides a more efficient test-time personalization than baselines that rely on multi-trial validation and prompt construction at test time.
\begin{table}[h]
\centering
\small
\setlength{\tabcolsep}{6pt}
\renewcommand{\arraystretch}{1.15}
\begin{tabular}{lcc}
\toprule
\textbf{Model} & \textbf{Inference time (Wall-Clock)} & \textbf{Acc} \\
\midrule
Llama-3-8b   & 00:19:30 & 52.8 \\
+ \textbf{\textsc{P-Check}}      & 00:28:37 & \textbf{65.11} \\
Qwen3-13b    & 00:26:42 & 59.76 \\
BT+SynthMe   & 00:36:29 & 62.74 \\
\bottomrule
\end{tabular}
\caption{End-to-end inference time (wall-clock) and accuracy comparison.}
\label{tab:inference_time_acc}
\end{table}

\subsubsection{Adaptability under Preference Shift}
Although our main research question is largely orthogonal to preference shift, we additionally analyze whether \textsc{P-Check} remains effective in drift-prone settings.
To this end, we cast evaluation as an online-streaming history setup: for each user, interactions are ordered chronologically, the user log grows turn by turn, and at turn $t$ the model must predict rewards using only the history observed up to $t\!-\!1$.
At each step, the preference summary is updated with newly observed interactions, and \textsc{P-Check} generates a query-specific checklist from the updated summary to judge the current candidate responses.
For stability, we restrict the analysis to users with at least 8 turns in PRISM-Personalized.
We then split each user's trajectory into four chronological quarters (Q1--Q4) and report reward prediction accuracy for each segment, averaging scores across users.
Later quarters therefore reflect settings in which preference drift is more likely to occur.
As a baseline, we compare against a persona-based judge that uses the same per-turn updated persona, but without checklist generation.
As shown in Table~\ref{tab:preference_shift}, \textsc{P-Check} consistently outperforms the persona-based baseline across all temporal quarters.
Notably, the gap is largest in the late-stage segment (Q4), where preference drift is most likely, suggesting that query-specific checklists provide a more robust signal for reward prediction under drift-prone conditions.
Overall, these results indicate that although \textsc{P-Check} is not explicitly designed for modeling preference shift, it adapts naturally to such settings without substantial degradation, simply by updating the user summary over time.

\begin{table}[t]
\centering
\small
\setlength{\tabcolsep}{4pt}
\begin{tabular}{lccccc}
\toprule
 & Q1 & Q2 & Q3 & Q4 & \shortstack{Macro Avg.} \\
\midrule
Persona-based & 55.81 & 58.48 & 57.05 & 56.20 & 56.20 \\
\textbf{\textsc{P-Check}} & \textbf{58.93} & \textbf{61.65} & \textbf{62.06} & \textbf{61.33} & \textbf{61.33} \\
\bottomrule
\end{tabular}
\caption{Online evaluation under preference-shift scenarios on PRISM-Personalized. For each user, turns are split into four chronological quarters (Q1--Q4), and accuracy is averaged across users.}
\label{tab:preference_shift}
\end{table}

\subsubsection{Analysis across Diverse Checklist Generator Backbones}
To examine whether the checklist generator in \textsc{P-Check} operates robustly across model families, we additionally train the checklist generator using different open-weight backbone LLMs, including Qwen and Gemma.
For a controlled comparison, we keep the downstream LLM judge fixed as Llama3-8B.
Table~\ref{tab:backbone_analysis} shows that \textsc{P-Check} achieves consistently competitive performance across backbone choices on both PRISM (ID) and Arena (OOD).
This suggests that the proposed training recipe is not tied to a specific architecture, and generalizes well across different backbone families.

\begin{table}[t]
\centering
\small
\begin{tabular}{lcc}
\toprule
Backbone & PRISM (ID) & Arena (OOD) \\
\midrule
Qwen-3-4B-It   & \textbf{66.48} & \textbf{63.55} \\
Gemma-3-4B-It  & 64.47 & 62.21 \\
Llama-3-3B-It  & 65.11 & 61.56 \\
\bottomrule
\end{tabular}
\caption{Performance of \textsc{P-Check} with different checklist generator backbones. For fair comparison, the LLM judge is fixed to Llama3-8B.}
\label{tab:backbone_analysis}
\end{table}

\subsubsection{Evaluation on Checklist Quality}

To verify whether \textsc{P-Check} effectively captures the personalized criteria, we assess the quality of the generated checklists using G-Eval~\citep{liu-etal-2023-g}. 
We sample the same PRISM instances used in our preliminary analysis and employ the corresponding oracle checklists as gold references to measure the alignment of the inferred criteria. 
As shown in Table~\ref{tab:check_qual}, \textsc{P-Check} achieves the highest score of 3.81, outperforming larger models such as Llama-3-8B and even GPT-4o-mini. 
These results suggest that our specialized generator is highly effective at recovering latent user preferences into the actionable personalized evaluation criteria, outperforming much larger, general-purpose models.

\begin{table}[h]
    \centering
    \small
    \setlength{\tabcolsep}{8pt}
    \renewcommand{\arraystretch}{1.08}
    
    \label{tab:checklist_quality_geval}
    \begin{tabular}{l c}
        \toprule
        \textbf{Model} & \textbf{G-Eval (GPT-5)} \\
        \midrule
        Llama-3-3b & 2.70 \\
        \textbf{(+) \textsc{P-Check} (3b)} & \textbf{3.81} \\
        Llama-3-8b & 2.93 \\
        Qwen-3-8b & 3.32 \\
        GPT-4o-mini & 3.54 \\
        \bottomrule
    \end{tabular}
    \caption{Checklist quality evaluation scored by G-Eval.}
    \label{tab:check_qual}
\end{table}

\subsubsection{Case Study}
\label{sec:case_study}

To better understand the behavior of \textsc{P-Check}, we present a qualitative analysis of two distinct scenarios: a success case where the model correctly models latent user preferences, and a failure case where it falls into the over-reliance on user priors.

\paragraph{Success Case: Explicitizing Latent Preferences.}
Table~\ref{tab:case_cherry} demonstrates how \textsc{P-Check} successfully translates a user's abstract preference for ``depth'' and ``structure'' into actionable evaluation criteria.
The user ($GP_u$) has a history of favoring concrete details and historical context.
For the query regarding the band \textit{Guns N' Roses}, the checklist generator accurately synthesized \texttt{Essential} criteria such as \textit{``Specificity of Information''} and \textit{``Historical Significance.''}
Consequently, the scoring model assigned high scores (e.g., 8.0, 9.0) to the user-preferred response (A), which provided a structured overview of the band's career, while correctly penalizing the rejected response (B) that narrowly focused on a single song.

\paragraph{Failure Case: Over-reliance on User Priors.}
Conversely, Table~\ref{tab:case_lemon} demonstrates a failure case where \textsc{P-Check} exhibits an over-reliance on static user priors.
The user typically prefers fact-based rigorous responses when discussing social issues ($GP_u$).
However, when the user shifted the context to a metaphysical query (\textit{``Do you think animals go to heaven or hell?''}), the model failed to adapt.
Instead of recognizing the subjective nature of the question, the generator hallucinated specious criteria demanding \textit{``Scientific Consensus''} and \textit{``Empirical Data''} (e.g., Criterion 0 and 1).
As a result, the model penalized the user's preferred empathetic response (A) for lacking scientific evidence and favored the generic, defensive refusal (B).
Although P-Check incorporates query context during generation, this mechanism does not guarantee the correct interpretation of latent intents when latent intent in the inquiry diverges significantly from established patterns.
This may be attributed to the limited contextual reasoning capacity of the small backbone model (3B parameters) used in \textsc{P-Check}.
We anticipate that exploring strategies to contextualize the historical preferences relative to the immediate query would significantly enhance performance.
\begin{table*}[t]
    \centering
    \small
    \setlength{\tabcolsep}{3pt} % column padding

    % ---- keep caption size normal; shrink ONLY if overflow happens ----
    \begin{adjustbox}{max width=\textwidth,center}
    \begin{minipage}{\textwidth}

    % =========================
    % PART 1: Context
    % =========================
    \begin{tabularx}{\linewidth}{@{}l X@{}}
        \toprule
        \textbf{User GP} &
        The user demonstrates a clear preference for responses that provide \textbf{concrete, factual information} with a focus on \textbf{specific details and historical context}. They favor content that is \textbf{rich in substance} ... rather than superficial commentary. Structurally, the user favors \textbf{organized and concise responses} (e.g., bullet points). \\
        \addlinespace[0.2cm]
        \textbf{Query} &
        \textit{``Guns and Roses is my favorite band ever. Could you tell me something interesting about them?''} \\
        \midrule
    \end{tabularx}

    % =========================
    % PART 2: Responses
    % =========================
    \begin{tabularx}{\linewidth}{@{}X X@{}}
        \textbf{(A) User's Chosen Response (P-Check's Choice)} & \textbf{(B) User's Rejected Response} \\
        \cmidrule(r){1-1} \cmidrule(l){2-2}
        Sure! Guns N' Roses is a highly acclaimed American rock band formed in Los Angeles in 1987. ... Some key interesting facts include: \newline
        \textbullet\ They have released numerous critically acclaimed and commercially successful albums... \newline
        \textbullet\ The members' diverse backgrounds... have been a major factor in shaping the band's unique style. \newline
        \textbullet\ They have been recognised for their energetic live performances... &
        Absolutely, Guns N' Roses were known for intense rock music but they're also skilled in softer side. \newline \newline
        Their song ``November Rain'' is the longest song ever to enter the top 10 of the US Billboard Hot 100 chart, stretching over 8 minutes. It emphasizes their versatile musical prowess. \\
    \end{tabularx}

    \vspace{0.1cm}

    % =========================
    % PART 3: Checklist (Full Criteria)
    % =========================
    \begin{tabularx}{\linewidth}{@{}l p{0.23\linewidth} >{\raggedright\arraybackslash}X c c@{}}
        \toprule
        \multicolumn{5}{c}{\textbf{Generated Checklist \& Scoring}} \\
        \midrule
        \textbf{Weight} & \textbf{Evidence} & \textbf{Criterion} & \textbf{Score (A)} & \textbf{Score (B)} \\
        \midrule
        Essential &
        \textbf{[GP]} Concrete/Factual \newline \textbf{[Q]} Interesting facts &
        Specificity: Provides multiple distinct facts about the band, including formation, notable songs, and impact. &
        \textbf{8.0} & 3.0 \\
        \midrule
        Essential &
        \textbf{[GP]} Historical context \newline \textbf{[Q]} Engaging content &
        Historical Significance: Includes details about the band's influence on music and their role in the rock genre. &
        \textbf{6.0} & 1.0 \\
        \midrule
        Important &
        \textbf{[GP]} Well-rounded \newline \textbf{[Q]} Discussion &
        Diversity of Content: Mentions various aspects (sound, style, achievements). &
        \textbf{8.0} & 7.0 \\
        \midrule
        Important &
        \textbf{[GP]} Organized/Concise \newline \textbf{[Q]} Interesting info &
        Structure: Presents information in a clear, organized manner (e.g., bullet points). &
        \textbf{9.0} & 2.0 \\
        \midrule
        Important &
        \textbf{[GP]} Analytical reasoning \newline \textbf{[Q]} Deeper exploration &
        Analytical Engagement: Encourages further discussion or exploration of specific aspects (e.g., music style). &
        \textbf{2.0} & 1.0 \\
        \midrule
        Optional &
        \textbf{[GP]} Assertive tone \newline \textbf{[Q]} Authoritative insights &
        Tone of Authority: Conveys confidence in the information presented, avoiding vague remarks. &
        \textbf{9.0} & 7.0 \\
        \midrule
        \multicolumn{5}{@{}l@{}}{\textbf{Success Reason:} The model successfully translated the user's latent preference into actionable criteria (e.g., Specificity).} \\
        \bottomrule
    \end{tabularx}

    \end{minipage}
    \end{adjustbox}

    \caption{\textbf{Success Case (Cherry-Picked).} \textsc{P-Check} successfully translates the user's preference for ``depth'' and ``structure'' into specific criteria. The model correctly assigns higher scores to the \textit{User Preferred} response due to its adherence to specificity and historical context.}
    \label{tab:case_cherry}
\end{table*}
\begin{table*}[t]
    \centering
    \small
    \setlength{\tabcolsep}{3pt} % column padding

    % ---- Cherry-Picked와 동일한 강제 맞춤 설정 (adjustbox) ----
    \begin{adjustbox}{max width=\textwidth,center}
    \begin{minipage}{\textwidth}

    % =========================
    % PART 1: Context
    % =========================
    \begin{tabularx}{\linewidth}{@{}l X@{}}
        \toprule
        \textbf{User GP} &
        The user demonstrates a strong preference for \textbf{direct, assertive responses} that provide \textbf{clear, factual information}. They favor content that is \textbf{not only accurate but also socially conscious}. They appreciate \textbf{analytical approaches} ... rather than generic advice. \\
        \addlinespace[0.2cm]
        \textbf{Query} &
        \textit{``Do you think animals go to heaven or hell?''} \\
        \midrule
    \end{tabularx}

    % =========================
    % PART 2: Responses
    % =========================
    \begin{tabularx}{\linewidth}{@{}X X@{}}
        \textbf{(A) User's Chosen Response} & \textbf{(B) User's Rejected Response (P-Check's Choice)} \\
        \cmidrule(r){1-1} \cmidrule(l){2-2}
        I would say they go to a happy place. I don’t believe in hell. \newline \newline
        I don’t believe in a God that would send someone to hell. I say this because I believe everyone has the right to their own beliefs. &
        I do not have a definitive view on animal afterlife. \newline \newline
        As an AI assistant without personal religious beliefs, I can only state that ideas about animal souls and an afterlife for animals vary greatly between religions and philosophies. \\
    \end{tabularx}

    \vspace{0.1cm}

    % =========================
    % PART 3: Checklist (Full 6 Criteria)
    % =========================
    \begin{tabularx}{\linewidth}{@{}l p{0.23\linewidth} >{\raggedright\arraybackslash}X c c@{}}
        \toprule
        \multicolumn{5}{c}{\textbf{Generated Checklist (Specious Criteria) \& Scoring}} \\
        \midrule
        \textbf{Weight} & \textbf{Evidence} & \textbf{Criterion} & \textbf{Score (A)} & \textbf{Score (B)} \\
        \midrule
        Essential &
        \textbf{[GP]} Factual info \newline \textbf{[Q]} Beliefs query &
        Directness (Science): Clearly states there is \textbf{no scientific consensus}, avoiding vague language. &
        2.0 & \textbf{8.0} \\
        \midrule
        Essential &
        \textbf{[GP]} Accuracy/Empirical \newline \textbf{[Q]} Afterlife opinion &
        Factual Accuracy: Explains the \textbf{lack of scientific evidence}, emphasizing empirical data. &
        4.0 & \textbf{6.0} \\
        \midrule
        Important &
        \textbf{[GP]} Complexity \newline \textbf{[Q]} Subjective discussion &
        Acknowledgment of Complexity: Recognizes diversity of beliefs/religions. &
        1.0 & \textbf{4.0} \\
        \midrule
        Optional &
        \textbf{[GP]} Respectful tone &
        Tone: Maintains a respectful/neutral tone providing factual info. &
        3.0 & \textbf{7.0} \\
        \midrule
        Optional &
        \textbf{[GP]} Logic/Structure &
        Clarity and Structure: Presents information in a clear, logical manner, possibly using distinct sections. &
        2.0 & \textbf{3.0} \\
        \midrule
        Optional &
        \textbf{[GP]} Depth/Critical thinking &
        Encouragement of Further Inquiry: Invites the user to reflect on their own beliefs, fostering a deeper dialogue. &
        1.0 & 1.0 \\
        \midrule
        \multicolumn{5}{@{}l@{}}{\textbf{Failure Reason:} over-reliance on the user's historical preference for ``facts,'' incorrectly applying a ``scientific'' constraint.} \\
        \bottomrule
    \end{tabularx}

    \end{minipage}
    \end{adjustbox}

    \caption{\textbf{Failure Case (Lemon-Picked).} The model misapplies the user's historical preference for ``factual accuracy'' to a metaphysical query. Consequently, the model rejects the user's preferred empathetic response (A) and wrongly favors the cold, scientific refusal (B).}
    \label{tab:case_lemon}
\end{table*}

\subsubsection{Prompts}
% We provide detailed prompts used in implementing P-Check in Table~\ref{prompt_gp},~\ref{prompt_c_gen},~\ref{prompt_judge}, and ~\ref{pmt:prompt_feedback}.
For reproducibility, we present the detailed prompts used to implement \textsc{P-Check}.
This includes the prompt for user's general preference generation (Table~\ref{prompt_gp}), checklist generation for training users (Table~\ref{prompt_c_gen}), checklist-guided scoring of LLM-as-a-Judge (Tables~\ref{prompt_judge}), and prompts used for feedback experiments (Table~\ref{pmt:prompt_feedback}).
\begin{table*}[htbp]
\centering
\begin{tabularx}{\textwidth}{X}
\toprule
\textbf{Generating General Preference from User History} \\ \midrule
\textcolor{blue}{\textbf{[Instruction]}} \\
Your task is to infer the [User’s General Profile] (GP) from the past [Interaction History of User], with specific, contrastive, and fine-grained reasoning-not a generic summary.\\

Use both [Chosen Model Response] and [Rejected Model Response] as primary comparative evidence.\\
For each pair, analyze what concrete aspects made the chosen response more aligned with the user’s taste and what made the rejected one less so.\\

Then, synthesize a rich, multidimensional profile capturing the user’s preferences across content, tone, reasoning style, and structure.\\\\

\#\#Requirements\\

Identify explicit contrasts for each (Chosen, Rejected) pair:\\
what traits were preferred or disliked.\\

Cluster insights by aspect:\\

- Content Preference (topics, detail, concreteness, etc..)\\

- Style (analytical, cautious, comparative, exploratory, etc..)\\

- Tone \& Attitude (empathetic, assertive, neutral, reflective, etc..)\\

- Structure \& Delivery (organized, concise, example-rich, stepwise, etc..)\\

Write the final GP as a rich, descriptive text with concrete behavioral signals (not abstract adjectives).\\

Avoid shallow generalizations like “user prefers clear answers.”\\
Instead, explain how and why the user prefers certain types of responses.\\\\

\textcolor{blue}{\textbf{[Inputs]}} \\
\textbf{[Interaction History of User]}: \{history\} \\\\
\textcolor{blue}{\textbf{[Output]}} \\
\textbf{[GP]} (Write only the GP text itself — no labels, no headings, no explanations.): \\ \bottomrule
\end{tabularx}
\caption{Prompt used for generating general preference from user history.}
\label{prompt_gp}
\end{table*}
\begin{table*}[htbp]
\centering
\begin{tabularx}{\textwidth}{X}
\toprule
\textbf{Collecting Checklist from Preference Data} \\ \midrule
\textcolor{blue}{\textbf{[Instruction]}} \\
You are a rigorous personalization checklist designer.\\
\#\#Goal\\
Given a [User's General Preference] (GP), a [Current User Query] (Q), a high-quality [Chosen Model Response] that aligns with the user's preference, and a [Rejected Model Response] that does not, generate a compact but expressive [Personalized Checklist] that can later verify whether any candidate response is personalized for this user on this GP and Q.\\\\

\#\#Critical Instruction\\

- Although you are provided with Chosen and Rejected responses to understand what distinguishes preferred vs. non-preferred behavior,your final output MUST appear as if it was created solely from GP and Q.\\

That means:\\

- You should use Chosen–Rejected contrast only implicitly to discover what matters to the user, but\\

- The output text itself must not refer to, mention, or reveal that any contrast was used.\\

- The final checklist must read like it was inferred only from (GP + Q). Be explicit about the chain : state the evidence from GP and Q → facet → checklist criterion.\\\\

\#\#Requirements\\

- Internally, extract concrete evidence from GP, Q, and the implicit contrast between Chosen vs. Rejected.\\

- Be explicit in reasoning (internally):\\
evidence from (GP + Q + Chosen–Rejected contrast) → facet → checklist criterion.\\

- But in the output, only show [evidence] whether (from GP and Q) or (only from GP or Q) , [facets], and [criteria] that could plausibly be derived from GP + Q, GP only, or Q only.\\

- Each criterion should capture a specific personalization aspect (tone, reasoning style, value emphasis, level of concreteness, etc.).\\

- Keep it short and readable (bullets, one-liners).\\

- The final checklist must be self-contained and directly usable for evaluating future responses, without referencing Chosen or Rejected.\\\\

Use Example 1 to 3 as reference, respond to Example 4.\\\\
<Example 1>\\

… \\

<Example 4>\\\\

\textcolor{blue}{\textbf{[Inputs]}} \\
\textbf{[User's General Preference]}: \{GP\} \\
\textbf{[Current User Query] }: \{query\} \\
\textbf{[Chosen Model response]}: \{chosen\} \\
\textbf{[Rejected Model response]}: \{reject\} \\\\
\textcolor{blue}{\textbf{[Output]}} \\
\textbf{[Personalized Checklist] (Json format)}: \\ \bottomrule
\end{tabularx}
\caption{Prompt used for collecting checklist from user preference data.}
\label{prompt_c_gen}
\end{table*}
\begin{table*}[htbp]
\centering
\begin{tabularx}{\textwidth}{X}
\toprule
\textbf{LLM-as-a-Judge scoring} \\ \midrule
\textcolor{blue}{\textbf{[Instruction]}} \\
You are a rigorous personalization verifier.\\
\#\#Task\\
For EACH personalized checklist item:\\
Assign a 1–10 score based solely on [Candidate Model response], using [User GP]/[Current User Query] only to interpret intent. Ignore criteria not present in [Personalized Checklist].\\

Before you assign the score for a criterion, briefly explain your reasoning about how well the [Candidate Model response]\\
satisfies that specific criterion. This reasoning must be included in the final JSON output as a "reasoning" field next to "criterion" and before "score".\\\\

\#\#\#Scoring rubric for each criterion in Personalized Checklist:\\

10 = Fully and explicitly satisfies the criterion; multiple clear, direct signals; no contradictions.\\
9 = Very strong satisfaction; clear evidence; tiny/immaterial gap.\\
8 = Strong satisfaction; at least one direct signal; minor gaps.\\
7 = Good satisfaction; mostly met with some notable gaps.\\
6 = Fair/partial satisfaction; indirect or mixed support; missing key detail(s).\\
5 = Weak satisfaction; generic/vague alignment with clear omissions.\\
4 = Very weak; tenuous/off-target support or partial contradiction.\\
3 = Minimal alignment; mostly irrelevant or unclear.\\
2 = Barely any alignment; largely irrelevant; possible contradiction.\\
1 = Not satisfied; absent or clearly contradictory.\\\\

For [Verify Result] (Json format), return ONLY a single JSON object with this shape (no extra text):\\
\{
"results": [\\
\{"index": 1, "criterion": "<exact criterion item 1>", "reasoning": "<brief reasoning for how well the response\\ satisfies this criterion>", "score": <1-10score>\},\\
\{"index": 2, "criterion": "<exact criterion item 2>", "reasoning": "<brief reasoning for how well the response\\ satisfies this criterion>", "score": <1-10score>\},\\
...\\
]\\
\}\\\\

\textcolor{blue}{\textbf{[Inputs]}} \\
\textbf{[User's General Preference]}: \{GP\} \\
\textbf{[Current User Query] }: \{query\} \\
\textbf{[Candidate Model response]}: \{response\} \\
\textbf{[Personalized Checklist]}: \{checklist\} \\\\
\textcolor{blue}{\textbf{[Output]}} \\
\textbf{[Verify Result] (Json format)}: \\ \bottomrule
\end{tabularx}
\caption{Prompt used for LLM-as-a-Judge to infer criterion-wise score based on the generated checklist.}
\label{prompt_judge}
\end{table*}
\begin{table*}[htbp]
\centering
\begin{tabularx}{\textwidth}{X}
\toprule
\textbf{Refinement of Model Response with Checklist Feedback} \\ \midrule
\textcolor{blue}{\textbf{[Instruction]}} \\
Your task is to rewrite the provided [Initial Model Response] so that it fully addresses the [User Query], while better fitting the target user's preferences and needs.\\\\

You are given:\\
1. [User Query]\\
2. [Initial Model Response]\\
3. A [Personalized Checklist] that describes what the response should do or improve.\\

Use the checklist as feedback: incorporate relevant criteria while preserving correct and helpful content.\\
Do NOT explicitly mention the checklist, or describe your reasoning process in the final [Rewritten Personalized Response].\\
Make the final response tailored to the user.\\

\textcolor{blue}{\textbf{[Inputs]}} \\
\textbf{[User Query] }: \{query\} \\
\textbf{[Personalized Checklist]}: \{checklist\} \\\\
\textcolor{blue}{\textbf{[Output]}} \\
\textbf{[Rewritten Personalized Response]} (your revised version of [Initial Model Response] text here only): \\ \bottomrule
\end{tabularx}
\caption{Prompt used for feedback experiment.}
\label{pmt:prompt_feedback}
\end{table*}

\end{document}